\documentclass{ieeeaccess}
\usepackage{cite}
\usepackage{amsmath,amssymb,amsfonts}
\usepackage{algorithmic}
\usepackage{soul}
\sethlcolor{yellow}
\usepackage{graphicx}
\usepackage{textcomp}
\usepackage{hyperref}
\usepackage{pifont}
\usepackage{dirtree}
\usepackage{ltablex} 
\keepXColumns          
\usepackage{array}
\newcommand{\cmark}{\ding{51}}           %
\newcommand{\xmark}{\ding{55}}           
\newcommand{\dmark}{$\Delta$}            

\def\BibTeX{{\rm B\kern-.05em{\sc i\kern-.025em b}\kern-.08em
    T\kern-.1667em\lower.7ex\hbox{E}\kern-.125emX}}
\begin{document}
\history{Received 10 December 2025, accepted 31 December 2025, date of publication 5 January 2026, date of current version 9 January 2026.}
\doi{10.1109/ACCESS.2026.3651218}

\title{Agentic AI in Healthcare \& Medicine: A Seven-Dimensional Taxonomy for Empirical Evaluation of LLM-based Agents}

\author{\uppercase{Shubham Vatsal}\authorrefmark{1},
\uppercase{Harsh Dubey\authorrefmark{2} and Aditi Singh}\authorrefmark{3} (Senior Member, IEEE)
}
\address[1]{Department of Computer Science, New York University, CIMS, New York, USA (e-mail: sv2128@nyu.edu)}
\address[2]{Department of Computer Science, New York University, CIMS, New York, USA (e-mail: hd2225@nyu.edu)}
\address[3]{Department of Computer Science, Cleveland State University, Cleveland, USA (e-mail: a.singh22@csuohio.edu)}


\corresp{Corresponding author: Aditi Singh (e-mail: a.singh22@csuohio.edu).}

\begin{abstract}
Large Language Model (LLM)-based agents that plan, use tools and act has begun to shape healthcare and medicine. Reported studies demonstrate competence on various tasks ranging from EHR analysis and differential diagnosis to treatment planning and research workflows. 
Yet the literature largely consists of overviews which are either broad surveys or narrow dives into a single capability (e.g., memory, planning, reasoning), leaving healthcare work without a common frame. We address this by reviewing 49 studies using a seven-dimensional taxonomy: Cognitive Capabilities, Knowledge Management, Interaction Patterns, Adaptation \& Learning, Safety \& Ethics, Framework Typology and Core Tasks \& Subtasks with 29 operational sub-dimensions. Using explicit inclusion and exclusion criteria and a labeling rubric (\textit{Fully Implemented} \cmark, \textit{Partially Implemented} \dmark, \textit{Not Implemented} \xmark), we map each study to the taxonomy and report quantitative summaries of capability prevalence and co-occurrence patterns. Our empirical analysis surfaces clear asymmetries. For instance, the External Knowledge Integration sub-dimension under Knowledge Management is commonly realized ($\sim$76\% \cmark) whereas Event-Triggered Activation sub-dimenison under Interaction Patterns is largely absent ($\sim$92\% \xmark) and Drift Detection \& Mitigation sub-dimension under Adaptation \& Learning is rare ($\sim$98\% \xmark). Architecturally, Multi-Agent Design sub-dimension under Framework Typology is the dominant pattern ($\sim$82\% \cmark) while orchestration layers remain mostly partial. Across Core Tasks \& Subtasks, information centric capabilities lead e.g., Medical Question Answering \& Decision Support and Benchmarking \& Simulation, while action and discovery oriented areas such as Treatment Planning \& Prescription still show substantial gaps ($\sim$59\% \xmark). Together, these findings provide an empirical baseline indicating that current agents excel at retrieval-grounded advising but require stronger adaptation and compliance platforms to move from early-stage systems to dependable systems.
\end{abstract}


\titlepgskip=-15pt

\maketitle

\section{Introduction}
\label{sec:introduction}


The advent of LLMs has significantly accelerated advancements across multiple domains, showcasing transformative potential far beyond traditional applications. These models, including GPT-4 \cite{achiam2023gpt} and PaLM \cite{chowdhery2023palm} are trained on extensive datasets comprising billions or even trillions of parameters enabling broad generalization and sophisticated understanding. Empirical studies have showcased that increasing model scale directly correlates with enhanced performance in complex reasoning, multi-step decision making and interactive planning tasks \cite{wei2022chain}. The enhanced cognitive capabilities of modern LLMs have rapidly driven their integration into critical sectors such as healthcare \cite{yang2023large}, medicine \cite{thirunavukarasu2023large}, finance \cite{li2023large} for risk assessment and investment strategies and education \cite{yan2024practical} for personalized learning and instructional assistance. Recent research has further evolved to emphasize prompt engineering which is a technique involving precise manipulation of natural language inputs to extract and enhance task specific reasoning capabilities of LLMs \cite{schulhoff2024prompt, gu2023systematic, sahoo2024systematic, vatsal2024survey, vatsal2025multilingual}. This shift underscores a critical progression in LLM research highlighting the transition from predictive modeling to sophisticated, context sensitive cognitive interactions. Furthermore, the emergence of LLM-based agents has marked another significant milestone enabling autonomous decision making and interactive task execution across diverse application areas \cite{xi2025rise}.


In healthcare and medicine, LLMs have demonstrated considerable promise in streamlining clinical workflows, enhancing diagnostic accuracy and supporting clinical research. For instance, recent studies have utilized LLMs to automate clinical documentation, significantly reducing clinician workload and improving the quality of clinical records \cite{leong2024efficient, baker2024chatgpt}. Similarly, LLM-based decision support systems have been applied to medical question answering, achieving expert level performance in interpreting complex medical literature and clinical guidelines \cite{singhal2023large}. Complementing these advances, LLMs achieve state-of-the-art results on biomedical machine reading comprehension benchmarks \cite{vatsal2024can} and show measurable gains on guideline-based prior-authorization question answering over noisy, real-world records \cite{vatsal2024canprior}. The integration of multimodal capabilities into LLMs enabling the interpretation of radiology scans and clinical images has further advanced diagnostic efficiency as has been seen in frameworks like ChatCAD \cite{wang2024interactive}. Apart from diagnostics, LLMs have been explored for enhancing treatment planning and supporting evidence-based medicine by synthesizing vast corpora of medical research and clinical guidelines \cite{nori2023capabilities}. In drug discovery, LLMs accelerate the identification of novel drug candidates and streamline clinical trial design which has significantly helped in reducing both time and cost \cite{chakraborty2023artificial}. Patient interaction and clinical communication have also benefited from LLM-powered conversational agents thereby improving adherence to treatment plans, facilitating telemedicine and enabling remote patient monitoring \cite{subramanian2024enhancing}. Although these advancements demonstrate substantial impact, they also highlight the need for continuous evaluation of LLMs in healthcare to maintain high standards of reliability and interpretability.

Recently, the development of LLM-based agents which are often described under the broader umbrella of Agentic AI has represented an emerging paradigm that capitalizes on the advanced cognitive capabilities of LLMs. These agents integrate multiple cognitive modules including reasoning, planning and memory management allowing them to do sophisticated problem solving and context aware interactions across a variety of domains. Recent research such as Chain-of-Thought prompting \cite{wei2022chain} and Tree-of-Thoughts reasoning \cite{yao2023tree} has highlighted the importance of modular cognitive architectures which help in facilitating the effective execution of complex multi-step reasoning tasks within Agentic AI design space. Additionally, memory-augmented architectures that incorporate episodic and semantic memory systems have further enhanced agents’ capacities to maintain continuity across interactions improving both task accuracy and reliability. Adaptive learning mechanisms including reinforcement learning with human feedback (RLHF) enable these agents to iteratively refine their behavior by integrating user input and environmental signals resulting in dynamic adaptation and alignment with evolving user objectives. As the landscape of LLM-based agents evolves, their impact in decision making is evident across healthcare, finance, law and education. This evolution is reshaping industry practices enabling more intelligent, adaptive, context aware systems that redefine how tasks are automated and decisions made.

Building on these foundational advancements, LLM-based agents have emerged as a pivotal innovation within healthcare and medicine. These agents integrate advanced reasoning and planning which help in facilitating complex tasks such as differential diagnosis and real-time patient monitoring. Research has shown their utility in automating critical workflows including patient triage \cite{lu2024triageagent}, laboratory result interpretation and EHR management \cite{wang2025colacare}. As Agentic AI systems, LLM-based agents show promise in augmenting clinical decision making through interactive dialogues that foster collaboration between medical professionals and AI systems. These agents also facilitate continuous medical education by providing real-time access to updated clinical knowledge, best practices and evidence-based guidelines. In specialized fields such as oncology, cardiology and surgery, LLM-based agents are being employed to support complex risk assessments and treatment optimization \cite{zhou2024zodiac, li2024potential}. As their integration deepens across healthcare, Agentic AI-driven LLM-based agents can assist in not only streamlining decision support but also enable more holistic care pathways.

\begin{figure*}
    \centering
    \includegraphics[width=0.8\linewidth]{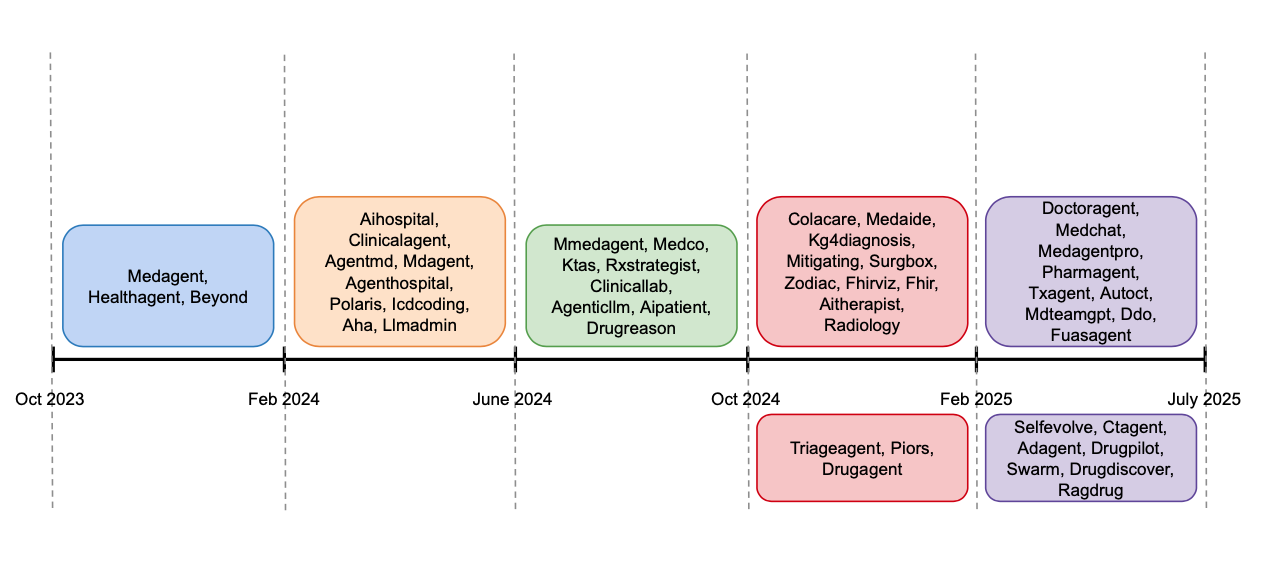}
    \caption{Timeline Diagram of All the Papers Evaluated in Our Work. Refer to Table \ref{tab:cognitive} for Mapping Between Paper Names and Corresponding Citations}
    \label{fig:timeline}
\end{figure*}

\normalsize
\subsection{Focus of Our Survey}

The growing interest in LLM-based agents has prompted exploration of potential applications in healthcare and medicine. Yet, the field lacks a structured framework for evaluating their multifaceted capabilities and system designs. Existing studies often examine specific applications or agent architectures in isolation, limiting a thorough understanding of their operational scope and performance dimensions. In this survey, we address this gap by conducting an extensive evaluation of 49 research papers on LLM-based agents deployed within healthcare and medicine. We propose an integrated taxonomy spanning 7 core dimensions and 29 sub-dimensions, systematically characterizing these agents across Cognitive Capabilities, Knowledge Management, Interaction Patterns, Adaptation \& Learning, Safety \& Ethics, Framework Typology and Core Tasks \& Subtasks. Our survey not only defines and contextualizes each dimension but also maps the selected studies against this taxonomy enabling a large scale quantitative analysis of prevailing methodologies. To aid synthesis, we include a taxonomy diagram that visualizes how the 49 studies map across the 7 dimensions and 29 sub-dimensions. We also derive broader insights that illuminate how different dimensions and sub-dimensions manifest in practice. These insights help us in emphasizing on diverse agent capabilities, domain specific methodologies and varying implementation approaches in healthcare and medicine. This extensive analysis provides a structured basis for developing robust, reliable and ethical Agentic AI-driven LLM agents in healthcare.


While several existing surveys \cite{guo2024large, ferrag2025llm, dong2024survey, li2024survey, plaat2025agentic, zhao2023depth} offer broad overviews of LLM-based agents, they tend to emphasize general purpose capabilities without delving deeply into healthcare and medicine specific applications. \cite{wang2025survey} introduces pertinent discussions around LLM-based agents in medicine but offers a limited exploration, covering roughly 25 papers. In contrast, our survey expands the scope significantly by analyzing 49 studies to capture a wider range of trends and advancements within healthcare and medicine. Similarly, surveys such as \cite{yu2025survey, zhang2404survey, li2025review, huang2024understanding, yehudai2025survey} examine select capabilities like memory, planning and reasoning in isolation. However, these fragmented perspectives fall short of providing a unified, multidimensional framework. Our work addresses this gap by delivering a comprehensive evaluation across 7 core dimensions and 29 sub-dimensions. This evaluation offers an integrated understanding that aligns with the unique challenges of healthcare and medicine.

Throughout this survey, we focus on a specific slice of Agentic AI: systems built around an LLM that drives the agent’s planning and decision-making, and can use tools, maintain memory or take actions. We do not cover agents that rely purely on symbolic reasoning, hand-crafted rules or architectures where the LLM is not the primary controller. Hybrid or multimodal setups are included only when the LLM is clearly the component coordinating the agent’s behavior.

\normalsize
\subsection{Literature Search Process}
We adopted a structured three-step filtering process to curate the research papers reviewed in this survey.


Step 1: We used multiple resources such as Google Scholar, PubMed, DBLP, Scopus and ArXiv to conduct a comprehensive search. Some of the search queries we used included \textit{LLM-based Agents in Healthcare}, \textit{LLM-based Agents in Medicine}, \textit{Using LLM-based Agents for Healthcare Tasks}, \textit{LLM-based Agents for Electronic Health Record Management}, \textit{LLM-based Agents for Drug Discovery}, \textit{LLM-based Agents for Clinical Documentation}, \textit{Using LLM-based Agents for Clinical Summarization} and \textit{LLM-based Agents in the Medical Domain}. This helped us gather an initial pool of research articles focused on LLM-based agents in healthcare and medicine. The initial search resulted in 137 articles after preliminary manual filtering to exclude irrelevant works. The preliminary manual screening removed papers for reasons such as: (i) duplicate postings of the same work across ArXiv and conference proceedings (ii) short workshop abstracts or extended abstracts lacking an implemented system (iii) papers whose contributions centered solely on dataset creation rather than agent behavior (iv) incomplete manuscripts including posters, project pages or early-stage preprints without technical detail. A group of 3 reviewers independently screened the initial pool of papers. Each paper was reviewed by at least 2 reviewers and disagreements were resolved in joint meetings through consensus. This process helped maintain consistency and minimize individual reviewer bias during inclusion/exclusion decisions.

    

    



Step 2: We applied two stringent selection criteria to further narrow the corpus. First, we included only studies in which an LLM is the core component of the agent’s design aligning with our focus on LLM-based agents. Second, we retained papers that explicitly address tasks in healthcare or medicine, while excluding studies primarily rooted in other domains. Because healthcare and medicine sometimes intersect with closely related fields such as biomedicine, we evaluated such overlaps case by case. If the core contribution of the paper was predominantly aligned with healthcare or medicine even in the presence of interdisciplinary elements, we included it in our survey.

Step 3: We then limited the corpus to articles published between October 2023 and June 2025. Figure \ref{fig:timeline} shows the timeline of all included papers. The combination of Steps 2 and 3 output gave us a curated set of 49 research papers that underpin our evaluation framework.

\begin{figure}
    \centering
    \includegraphics[width=1.4\linewidth]{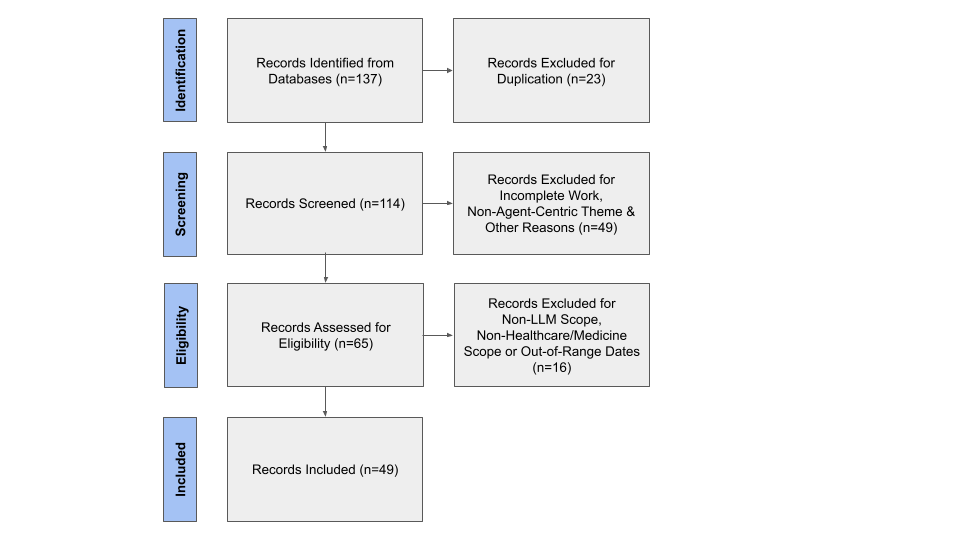}
    \caption{PRISMA Flow Diagram of the Study Selection Process}
    \label{fig:prisma}
\end{figure}

To summarize the full pipeline, we also include a PRISMA-style flow diagram in Figure \ref{fig:prisma} outlining identification, screening, eligibility assessment and final inclusion of the 49 studies in this survey.

\normalsize
\subsection{Outline}

The remainder of the paper is structured as follows. Section 2 introduces the 7 core dimensions and 29 sub-dimensions that form the basis of our evaluation framework, providing detailed definitions and motivations for each. In Section 3, we present key insights derived from our analysis, including patterns such as the most frequently employed sub-dimensions in healthcare and medicine, methodological trends and dominant architectural choices. Finally, Section 4 concludes the paper by summarizing our contributions.

\normalsize
\section{Evaluation Dimensions}

This section introduces the 7 evaluation dimensions and 29 sub-dimensions that structure our analysis of 49 studies. These dimensions are derived through a mixed methodology combining inductive extraction, structured synthesis and cross-referencing with prior frameworks. We began by performing open coding on all 49 papers, identifying recurring mechanisms related to planning, memory, interaction, learning, safety and system structure. These codes are then iteratively grouped into higher-level constructs through constant comparison until stable themes emerged. In parallel, we consulted existing surveys on LLM-based agents, agentic architectures and healthcare-AI evaluation to ensure conceptual continuity while avoiding direct replication. Dimensions are retained only when: (i) they appeared in several independent papers (ii) they captured operational behaviors that could be scored consistently (iii) they are not already subsumed by a broader construct. Sub-dimensions are defined a priori only when the literature already used a well-established term (e.g., planning, knowledge integration). All remaining sub-dimensions are inductively extracted from empirical implementation patterns in the 49-paper corpus. This combined process ensures that the taxonomy is both grounded in literature and empirically reproducible.

We emphasize that these definitions are practical and designed for consistent scoring across papers. They are not meant to be exhaustive. Healthcare LLM agents are evolving quickly and accordingly scholars may draw these boundaries differently. Our aim is operational clarity and cross-study comparability and not completeness. Each dimension admits multiple valid readings and some capabilities can appear under more than one heading depending on the system design and reported evidence. For the same reason, the categorical labels we apply \textit{Fully Implemented} \cmark, \textit{Partially Implemented} \dmark\ and \textit{Not Implemented} \xmark\ should be read as decision aids rather than hard thresholds. They are not strictly mutually exclusive. Systems often straddle boundaries and overlap can reasonably occur between \cmark\ and \dmark\ or between \dmark\ and \xmark\ depending on deployment context and evidence granularity. To handle these gray areas consistently, we favor conservative assignment grounded in verifiable descriptions. We default to \dmark\ where claims are implicit which means they are based only on ablations or demonstrated solely in simulation. We use \xmark\ where a mechanism is asserted without concrete procedure or evaluation. We assign \cmark\ where the capability is implemented end-to-end with explicit procedures and demonstrations. Despite the inherent interpretive space, our goal is comparability. We therefore standardize terminology across papers, apply common decision rules within each sub-dimension and record concise rationales explaining how evidence maps to labels. This yields a uniform reading of heterogeneous designs spanning clinical tasks and data modalities while leaving room for nuance and evolution. The taxonomy and labels in this study should be treated as a transparent baseline for cross-study synthesis and not as prescriptive judgments. Subsequent sections detail the rubrics we used so that others can reproduce, refine or contest specific assignments as the field matures.

\begin{figure*}[!t]
  \centering
  \begin{minipage}{\linewidth}
    \centering
    \includegraphics[width=1\linewidth]{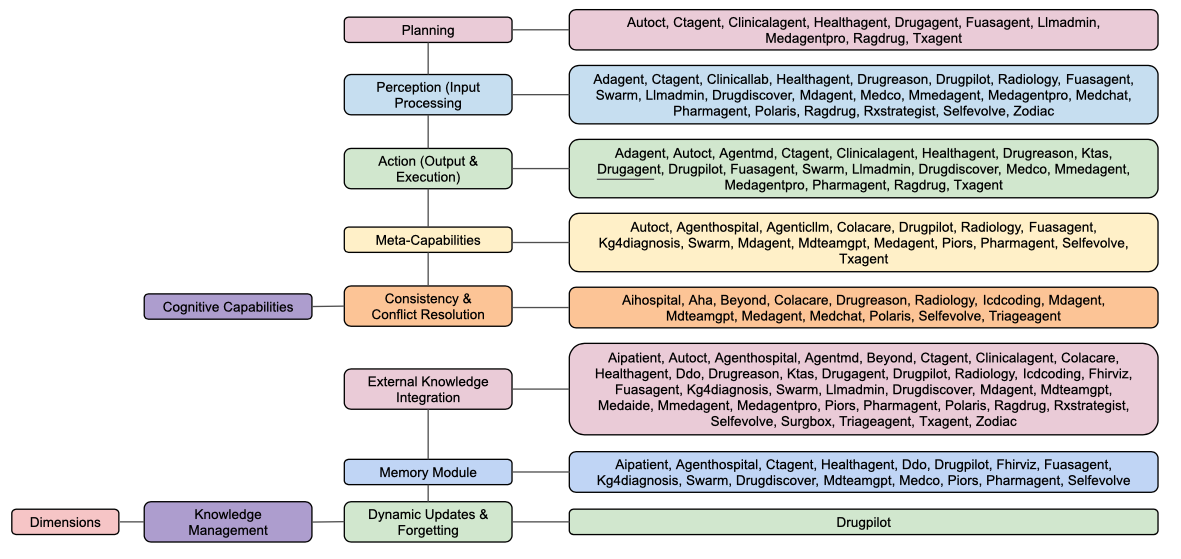}
  \end{minipage}\par\vspace{0.0001em}
  \begin{minipage}{\linewidth}
    \centering
    \includegraphics[width=1\linewidth]{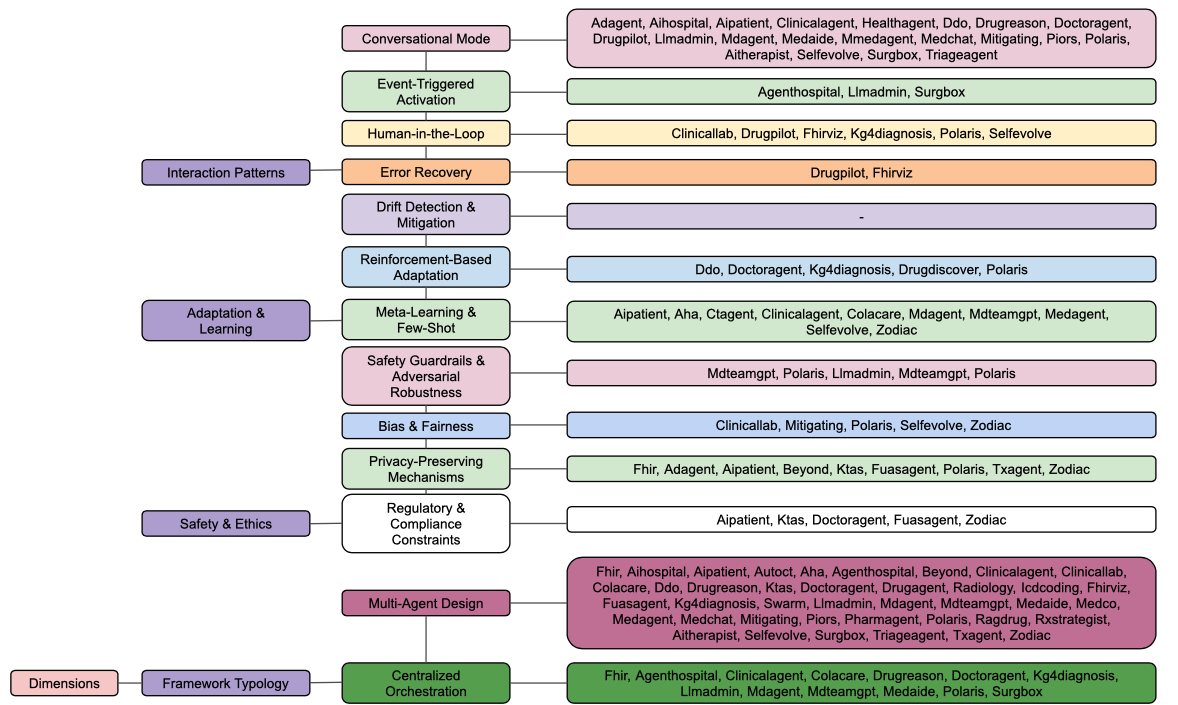}
  \end{minipage}

  \caption{Taxonomy Diagram of Dimensions Cognitive Capabilities, Knowledge Management, Interaction Patterns, Adaptation \& Learning, Safety \& Ethics, Framework Typology and their Corresponding Sub-Dimensions with Research Papers Rated \textit{Fully Implemented}. Refer to Table \ref{tab:cognitive} for Mapping Between Paper Names and Corresponding Citations.}
  \label{fig:tree_1}
\end{figure*}

\begin{figure*}[!t]
  \centering
  \begin{minipage}{\linewidth}
    \centering
    \includegraphics[width=1\linewidth]{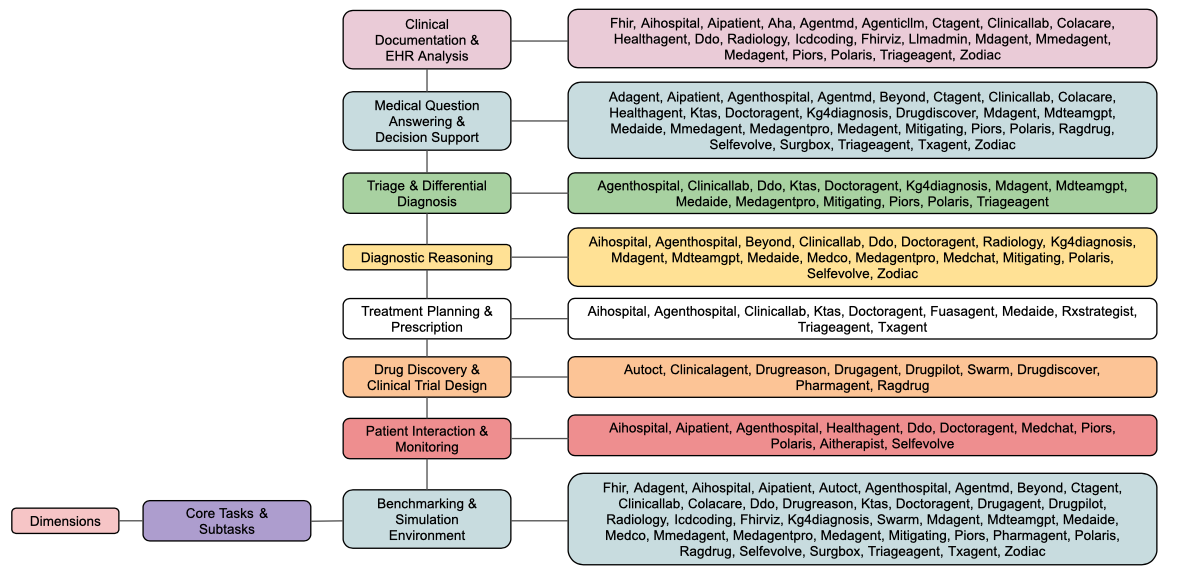}
  \end{minipage}\par\vspace{0.5em}
  \caption{Taxonomy Diagram of Dimension Core Tasks \& Subtasks and it's Corresponding Sub-Dimensions with Research Papers Rated \textit{Fully Implemented}. Refer to Table \ref{tab:cognitive} for Mapping Between Paper Names and Corresponding Citations.}
  \label{fig:tree_2}
\end{figure*}

The taxonomy diagram in Figure \ref{fig:tree_1} and \ref{fig:tree_2} visualizes our survey’s structure. In this diagram, the 7 evaluation dimensions are grouped with their 29 sub-dimensions. For each sub-dimension, we list every study rated \textit{Fully Implemented} \cmark\ under our rubric. This view acts as a navigational index. It highlights clusters of maturity (sub-dimensions with many \cmark\ papers), brings our attention to sparse areas where implementations are rare and surfaces studies with multi-capability designs. This diagram clarifies overlap among neighboring constructs and complements the tabular summaries. 


\normalsize
\subsection{Cognitive Capabilities}

\label{cognitive}

Agents must first translate clinical goals into workable courses of action, breaking long horizon objectives into milestones and decision points. Contemporary prompting strategies like Tree-of-Thoughts \cite{yao2023tree} and Plan-and-Solve \cite{wang2023plan} formalize this decomposition and search. In healthcare, such structured lookahead supports care pathway drafting, triage routing and prior-authorization checks under real-world constraints. These plans depend on semantic intake from heterogeneous inputs including notes, labs, images, device streams and guidelines. Multimodal pretraining shows how cross-modal alignment enables robust grounding \cite{alayrac2022flamingo} while domain tuned language models such as Med-PaLM 2 \cite{singhal2025toward} capture medical nomenclature and discourse. By converting raw EHR and imaging signals into clinically meaningful representations, agents can anchor plans to patient specific context rather than surface strings. Agents interleave reasoning with tool calls as exemplified in ReAct \cite{yao2023react} and learn when and how to invoke APIs as shown in Toolformer \cite{schick2023toolformer}. Because clinical stakes demand vigilance, agents benefit from self-monitoring that critiques intermediate products. Reflective control loops maintain episodic traces to improve future iterations \cite{shinn2023reflexion} while diversity aware decoding aggregates multiple rationales to stabilize conclusions \cite{wang2022self}. These mechanisms surface deferral conditions, steer escalation to humans and promote conservative behavior when evidence is weak. Finally, reliable clinical decisions require convergence across competing hypotheses and sources. Multi-agent deliberation frameworks coordinate debate and mediation toward consensus \cite{wu2024autogen} whereas sampling-based cross-checks expose factual divergence indicative of hallucination \cite{manakul2023selfcheckgpt}. Such protocols reconcile guideline recommendations with patient level data, reduce contradictory orders and justify suggestions with traceable agreement. Collectively, these capabilities constitute the Cognitive Capabilities of LLM-based agents. Table \ref{tab:cognitive} reports evaluations across the 5 Cognitive Capabilities sub-dimensions:


\normalsize
\subsubsection{Planning}

Planning is the agent’s capacity to synthesize, maintain and revise executable task structures that link objectives to actions. It entails principled task decomposition, temporally coherent sequencing and hierarchical organization that distinguishes strategic intent from tactical steps. Robust planning agents also enumerate alternative strategies, select among them and run critique-refine loops to repair omissions or dead ends. Within this survey, we judge Planning \cmark\ when there is clear evidence of dynamic plan construction with multi-step decomposition, explicit comparison of alternatives or refinement cycles and appropriate tool selection. We judge it \dmark\ when there is task breakdown or dynamic role assignment but no substantive strategy generation, comparison or iterative revision. It is \xmark\ when workflows are fixed or limited to simple conditional branching. In clinical contexts, the implemented form supports adaptive diagnostic workups, multi-step prior-authorization reviews and iterative treatment planning aligned to evolving patient data.

\normalsize
\subsubsection{Perception (Input Processing)}

Perception is the transformation of heterogeneous inputs into reliable, model usable representations through normalization, denoising and cross-modal fusion. Beyond tokenization, it encompasses various other things like ASR for speech, temporal alignment for waveforms, structure induction for EHR fields etc. This is followed by construction of provenance sensitive embeddings suitable for downstream reasoning and action. For evaluation, \cmark\ denotes robust pipelines that integrate multiple modalities via dedicated encoders and cross-modal fusion, exposing calibrated representations to the agent's planning and action components. \dmark\ describes systems that perform competent text parsing or per-modality encoding with minimal fusion and limited preprocessing. \xmark\ applies when inputs are passed as raw strings with no normalization or multimodal handling.

\normalsize
\subsubsection{Action (Output \& Execution)}

Action (Output \& Execution) is the agent’s execution layer that converts internal decisions into externally verifiable effects. It encompasses controlled text production along with tool and API invocation. Mature executors validate preconditions and postconditions, manage side effects and route observations back to the controller for closed loop operation. We label this sub-dimension \cmark\ when the system demonstrates actual tool use and external execution beyond text generation. It is \dmark\ when actions are limited to basic API calls or database lookups without richer tool orchestration or environment manipulation. It is \xmark\ when outputs are text only. In healthcare settings, the \cmark\ level enables safe medication reconciliation, order validation and prior-authorization checks with auditable traces.

\normalsize
\subsubsection{Meta-Capabilities}

Meta-Capabilities comprise introspective control over an agent’s own reasoning and actions. It comprises of generating critiques of intermediate steps, calibrating confidence, identifying knowledge gaps and deciding when to abstain or seek evidence. In clinical contexts, this layer enables agents to surface uncertainty in differential diagnoses, trigger literature retrieval or second opinions and defer to clinicians when evidence is insufficient. For evaluation, \cmark\ indicates explicit critique and revision loops, calibrated uncertainty estimates tied to deferral and logged escalation. \dmark\ reflects ad hoc self-monitoring like memory feedback or heuristic self-checks. \xmark\ indicates absence of self-critique, confidence signaling or deferral behavior.

\normalsize
\subsubsection{Consistency \& Conflict Resolution}

Consistency \& Conflict Resolution in agents refers to the systematic identification and reconciliation of contradictory information originating from parametric recall, external retrieval or multi-agent outputs. Effective implementations deploy layered strategies including contradiction detection through natural language inference models and ensemble agreement protocols. In complex domains like healthcare, such mechanisms are vital for ensuring that recommendations remain coherent, evidence-based and aligned with established clinical standards. The extent of realization can vary: (a) \textit{Fully Implemented} \cmark\ systems incorporate automated conflict detection with structured resolution pipelines using majority vote ensembling and adaptive re-weighting of specialist outputs (b) \textit{Partially Implemented} \dmark\ systems may rely on limited validation or ad hoc human oversight without comprehensive reconciliation (c) \textit{Not Implemented} \xmark\ omit explicit mechanisms altogether, increasing the risk of unsafe outputs.

\begin{table}
\caption{Evaluation of Cognitive Capabilities}

\setlength{\tabcolsep}{3pt}
\begin{tabular}{|p{60pt}|p{15pt}|p{35pt}|p{21pt}|p{25pt}|p{37pt}| }
\hline
\textbf{Papers} & \textbf{Plan} & \textbf{Perception}  & \textbf{Action} & \textbf{Meta Capabilities} & \textbf{Consistency \& Conflict Resolution} \\
\hline

Colacare \cite{wang2025colacare} & \xmark & \dmark & \xmark & \cmark & \cmark \\
    \hline
Medagent \cite{tang2023medagents} & \xmark & \dmark & \xmark & \cmark & \cmark\\
    \hline
Medaide \cite{yang2024medaide} & \dmark & \dmark & \xmark & \xmark & \xmark \\
    \hline
Doctoragent \cite{feng2025doctoragent} & \dmark & \dmark & \xmark & \xmark & \xmark \\
    \hline
Aihospital \cite{fan2024ai} & \xmark & \dmark & \xmark & \xmark & \cmark \\
    \hline
Mmedagent\cite{li2024mmedagent} & \dmark & \cmark & \cmark & \xmark & \xmark \\
    \hline
Medchat \cite{liu2025medchat} & \dmark & \cmark & \xmark & \xmark & \cmark \\
    \hline
Medco \cite{wei2024medco} & \xmark & \cmark & \cmark & \dmark & \xmark \\
    \hline
Medagentpro\cite{wang2025medagent} & \cmark & \cmark & \cmark & \dmark & \xmark \\
    \hline
Clinicalagent\cite{yue2024clinicalagent} & \cmark & \dmark & \cmark & \xmark & \xmark \\
    \hline
Ktas \cite{han2024development} & \xmark & \dmark & \cmark & \xmark & \xmark \\
    \hline
Agentmd \cite{jin2024agentmd} & \xmark & \dmark & \cmark & \xmark & \xmark\\
    \hline
Mdagent \cite{kim2024mdagents} & \dmark & \cmark & \xmark & \cmark & \cmark\\
    \hline
Agenthospital\cite{li2024agent} & \xmark & \dmark & \xmark & \cmark & \xmark\\
    \hline
Pharmagent \cite{gao2025pharmagents} & \xmark & \cmark & \cmark & \cmark & \xmark\\
    \hline
Txagent \cite{gao2025txagent} & \cmark & \dmark & \cmark & \cmark & \xmark\\
    \hline
Autoct \cite{liu2025autoct} & \cmark & \dmark & \cmark & \cmark & \xmark \\
    \hline
Kg4diagnosis\cite{zuo2025kg4diagnosis} & \dmark & \dmark & \xmark & \cmark & \xmark\\
    \hline
Mdteamgpt \cite{chen2025mdteamgpt} & \dmark & \dmark & \xmark & \cmark & \cmark\\
    \hline
Mitigating \cite{ke2024mitigating} & \xmark & \dmark & \xmark & \dmark & \dmark \\
    \hline
Polaris \cite{mukherjee2024polaris} & \dmark & \cmark & \xmark & \dmark & \cmark\\
    \hline
Rxstrategist \cite{van2024rx} & \xmark & \cmark & \xmark & \xmark & \xmark\\
    \hline
Clinicallab \cite{yan2024clinicallab} & \dmark & \cmark & \xmark & \xmark & \xmark\\
    \hline
Agenticllm \cite{sudarshan2024agentic} & \xmark & \dmark & \xmark & \cmark & \xmark\\
    \hline
Surgbox \cite{wu2024surgbox} & \dmark & \dmark & \xmark & \dmark & \xmark\\
    \hline
Aipatient \cite{yu2024simulated} & \xmark & \dmark & \xmark & \dmark & \dmark\\
    \hline
Healthagent \cite{abbasian2023conversational} & \cmark & \cmark & \cmark & \xmark & \xmark\\
    \hline
Icdcoding \cite{li2024exploring} & \xmark & \dmark & \xmark & \xmark & \cmark\\
    \hline
Aha \cite{pandey2024advancing} & \xmark & \dmark & \xmark & \xmark & \cmark\\
    \hline
Ddo \cite{jia2025ddo} & \dmark & \dmark & \xmark & \xmark & \xmark\\
    \hline
Beyond \cite{wang2024beyond} & \xmark & \dmark & \xmark & \xmark & \cmark\\
    \hline
Zodiac \cite{zhou2024zodiac} & \xmark & \cmark & \dmark & \xmark & \dmark\\
    \hline
Fhirviz \cite{almutairi2024fhirviz} & \xmark & \dmark & \dmark & \xmark & \xmark\\
    \hline
Fhir \cite{de2024multi} & \dmark & \dmark & \dmark & \xmark & \xmark\\
    \hline
Fuasagent \cite{zhao2025autonomous} & \cmark & \cmark & \cmark & \cmark & \xmark\\
    \hline
Aitherapist \cite{wasenmuller2024script} & \dmark & \dmark & \xmark & \xmark & \xmark\\
    \hline
Radiology \cite{zeng2024enhancing} & \xmark & \cmark & \xmark & \cmark & \cmark \\
    \hline
Triageagent \cite{lu2024triageagent} & \xmark & \dmark & \xmark & \xmark & \cmark\\
    \hline
Piors \cite{bao2024piors} & \dmark & \dmark & \dmark & \cmark & \xmark\\
    \hline
Selfevolve \cite{almansoori2025self} & \xmark & \cmark & \xmark & \cmark & \cmark\\
    \hline
Drugagent \cite{liu2024drugagent} & \cmark & \dmark & \cmark & \xmark & \xmark\\
    \hline
Ctagent \cite{mao2025ct} & \cmark & \cmark & \cmark & \xmark & \xmark\\
    \hline
Adagent \cite{hou2025adagent} & \dmark & \cmark & \cmark & \xmark & \dmark\\
    \hline
Drugpilot \cite{li2025drugpilot} & \dmark & \cmark & \cmark & \cmark & \xmark\\
    \hline
Llmadmin \cite{gebreab2024llm} & \cmark & \cmark & \cmark & \xmark & \xmark\\
    \hline
Swarm \cite{song2025llm} & \xmark & \cmark & \cmark & \cmark & \xmark\\
    \hline
Drugdiscover\cite{ock2025large} & \xmark & \cmark & \cmark & \dmark & \xmark\\
    \hline
Ragdrug \cite{lee2025rag} & \cmark & \cmark & \cmark & \xmark & \dmark\\
    \hline
Drugreason\cite{inoue2025drugagent} & \dmark & \cmark & \cmark & \xmark & \cmark\\
    \hline
\end{tabular}
\label{tab:cognitive}
\end{table}

\normalsize
\subsection{Knowledge Management}

\label{knowledge}

Knowledge Management in LLM-based agents centers on how parametric knowledge is complemented and routed through external resources when the model’s internal recall is insufficient. Agents arbitrate between weights and the world by invoking retrieval-augmented generation (RAG) \cite{lewis2020retrieval}, pretraining time retrieval such as REALM \cite{guu2020retrieval} and by interpolating with non-parametric memories as in kNN-LM \cite{khandelwal2019generalization}. In clinical settings, this blend helps surface up-to-date guidelines and trial evidence with citations for clinician review. Equally important is the agent’s internal memory architecture. Short-term buffers maintain discourse and tool use state. On the other hand, long-term episodic and semantic stores compress prior interactions and domain ontologies under learned controllers that page information into and out of context as in MemGPT \cite{packer2023memgpt}. These mechanisms reduce context length bottlenecks, stabilize multi-step plans and enable retrieval policies that are sensitive to patient specific history. In healthcare, such memory supports longitudinal summarization, medication reconciliation and cross-encounter continuity where sustained context improves care quality. Finally, knowledge must evolve and accordingly agents need to continuously index new documents, prioritize fresh sources and de-emphasize stale entries. Self-reflective retrieval frameworks that critique and re-query \cite{asai2024self} can improve factuality and citation reliability while targeted parametric edits can correct entrenched facts at scale \cite{meng2022mass}. In clinical deployments, such update and forget mechanisms are vital for aligning agent outputs with evolving guidelines and newly published trials. Table \ref{tab:knowledge} summarizes outcomes across its 3 sub-dimensions:

\normalsize
\subsubsection{External Knowledge Integration}

External Knowledge Integration refers to an agent’s capability to augment its intrinsic parametric knowledge encoded in model weights with non-parametric information. This non-parametric information is retrieved from external, often domain specific  repositories at inference time. Such repositories may include structured clinical databases, medical knowledge graphs, regulatory guideline repositories or unstructured corpora embedded in vector indices. The integration process requires not only the retrieval of relevant content but also mechanisms for arbitration between internally stored and externally sourced knowledge ensuring outputs are both temporally current and contextually precise. In healthcare and clinical decision support, this enables an LLM-based agent to ground its reasoning in authoritative sources such as drug interaction databases, treatment guidelines or up-to-date clinical trial results. The degree to which an agent embodies this sub-dimension can vary: (a) fully realized implementations \cmark\ incorporate structured RAG pipelines with calibrated arbitration between internal recall and external evidence (b)  partial implementations \dmark\ rely on limited, static or manually curated external content (c) absent implementations \xmark\ operate purely from parametric memory.

\begin{table}
\caption{Evaluation of Knowledge Management}
\setlength{\tabcolsep}{3pt}
\centering
\begin{tabular}{|p{60pt}|p{40pt}|p{35pt}|p{40pt}|}

\hline

\textbf{Papers} & \textbf{External Knowledge Integration} & \textbf{Memory Module}   & \textbf{Dynamic Updates \& Forgetting}\\
\hline

Colacare \cite{wang2025colacare} & \cmark & \dmark & \xmark\\
\hline
Medagent \cite{tang2023medagents} & \xmark & \xmark & \xmark\\
\hline
Medaide \cite{yang2024medaide} & \cmark & \dmark & \xmark \\
\hline
Doctoragent \cite{feng2025doctoragent} & \xmark & \dmark & \xmark  \\
\hline
Aihospital \cite{fan2024ai} & \xmark & \xmark & \xmark \\
\hline
Mmedagent \cite{li2024mmedagent} & \cmark & \dmark & \xmark \\
\hline
Medchat \cite{liu2025medchat} & \xmark & \xmark & \xmark \\
\hline
Medco \cite{wei2024medco} & \xmark & \cmark & \dmark \\
\hline
Medagentpro \cite{wang2025medagent} & \cmark & \dmark & \xmark \\
\hline
Clinicalagent \cite{yue2024clinicalagent} & \cmark & \dmark & \xmark \\
\hline
Ktas \cite{han2024development} & \cmark & \dmark & \xmark \\
\hline
Agentmd \cite{jin2024agentmd} & \cmark & \dmark & \xmark \\
\hline
Mdagent \cite{kim2024mdagents} & \cmark & \dmark & \xmark \\
\hline
Agenthospital \cite{li2024agent} & \cmark & \cmark & \dmark \\
\hline
Pharmagent \cite{gao2025pharmagents} & \cmark & \cmark & \dmark \\
\hline
Txagent \cite{gao2025txagent} & \cmark & \dmark & \xmark \\
\hline
Autoct \cite{liu2025autoct} & \cmark & \dmark & \xmark \\
\hline
Kg4diagnosis \cite{zuo2025kg4diagnosis} & \cmark & \cmark & \dmark \\
\hline
Mdteamgpt \cite{chen2025mdteamgpt} & \cmark & \cmark & \dmark \\
\hline
Mitigating \cite{ke2024mitigating} & \xmark & \xmark & \xmark \\
\hline
Polaris \cite{mukherjee2024polaris} & \cmark & \dmark & \xmark \\
\hline
Rxstrategist \cite{van2024rx} & \cmark & \dmark & \xmark \\
\hline
Clinicallab \cite{yan2024clinicallab} & \xmark & \xmark & \xmark \\
\hline
Agenticllm \cite{sudarshan2024agentic} & \xmark & \xmark & \xmark \\
\hline
Surgbox \cite{wu2024surgbox} & \cmark & \dmark & \xmark \\
\hline
Aipatient \cite{yu2024simulated} & \cmark & \cmark & \dmark \\
\hline
Healthagent \cite{abbasian2023conversational} & \cmark & \cmark & \dmark \\
\hline
Icdcoding \cite{li2024exploring} & \cmark & \dmark & \xmark \\
\hline
Aha \cite{pandey2024advancing} & \xmark & \xmark & \xmark \\
\hline
Ddo \cite{jia2025ddo} & \cmark & \cmark & \dmark \\
\hline
Beyond \cite{wang2024beyond} & \cmark & \dmark & \xmark \\
\hline
Zodiac \cite{zhou2024zodiac} & \cmark & \dmark & \xmark \\
\hline
Fhirviz \cite{almutairi2024fhirviz} & \cmark & \cmark & \dmark  \\
\hline
Fhir \cite{de2024multi} & \dmark & \dmark & \xmark  \\
\hline
Fuasagent \cite{zhao2025autonomous} & \cmark & \cmark & \dmark  \\
\hline
Aitherapist \cite{wasenmuller2024script} & \xmark & \xmark & \xmark  \\
\hline
Radiology \cite{zeng2024enhancing} & \cmark & \dmark & \xmark  \\
\hline
Triageagent \cite{lu2024triageagent} & \cmark & \dmark & \xmark  \\
\hline
Piors \cite{bao2024piors} & \cmark & \cmark & \dmark  \\
\hline
Selfevolve \cite{almansoori2025self} & \cmark & \cmark & \dmark  \\
\hline
Drugagent \cite{liu2024drugagent} & \cmark & \dmark & \xmark \\
\hline
Ctagent \cite{mao2025ct} & \cmark & \cmark & \dmark \\
\hline
Adagent \cite{hou2025adagent} & \xmark & \xmark & \xmark \\
\hline
Drugpilot \cite{li2025drugpilot} & \cmark & \cmark & \cmark \\
\hline
Llmadmin \cite{gebreab2024llm} & \cmark & \dmark & \xmark \\
\hline
Swarm \cite{song2025llm} & \cmark & \cmark & \dmark \\
\hline
Drugdiscover \cite{ock2025large} & \cmark & \cmark & \dmark \\
\hline
Ragdrug \cite{lee2025rag} & \cmark & \dmark & \xmark \\
\hline
Drugreason \cite{inoue2025drugagent} & \cmark & \dmark & \xmark\\
\hline
\end{tabular}
\label{tab:knowledge}
\end{table}

\normalsize
\subsubsection{Memory Module}

Memory Module in LLM-based agents provide structured mechanisms for storing and retrieving information beyond the transient context window enabling continuity and longitudinal reasoning. Short-term memory mechanisms only maintain immediate conversational history. However, long-term stores capture episodic narratives like past consultations and semantic abstractions such as knowledge graphs. These modules are often mediated by learned or rule-based controllers that determine what information should be committed to persistent storage and when historical records should be recalled for decision making. In healthcare applications, such architectures support richer patient interactions allowing agents to reference prior lab results, follow-up notes or evolving care plans without reintroduction by the user. Systems may fully implement \cmark\ this sub-dimension through sophisticated, persistent multi-tier memory with dynamic read/write control. Systems may partially implement \dmark\ it through limited short-term session scoped recall or basic retrieval buffers. Systems may not implement \xmark\ it at all relying solely on the model’s default context window.

\normalsize
\subsubsection{Dynamic Updates \& Forgetting}
Dynamic Updates \& Forgetting captures an agent’s ability to keep its knowledge current and hygienic rather than merely cumulative. Beyond simple retrieval, the system must: (i) ingest new evidence via continuous indexing (ii) re-weight stored items using temporal decay (iii) remove stale or retracted facts through explicit pruning rules. In clinical contexts, these mechanisms align outputs with evolving guidelines, drug recalls and newly published or retracted trials thus reducing the risk of obsolete recommendations. We consider this \cmark\ when both sides of knowledge maintenance are present: active addition (continuous updates/write-back) and explicit forgetting (time-decay thresholds and deprecation policies). It is \dmark\ when updates occur but no formal decay is defined resulting in static accumulation over time. It is \xmark\ when the system is static with no continuous updates decay or pruning beyond the model’s default context.

\normalsize
\subsection{Interaction Patterns}

Interaction patterns specify how an agent is invoked, sustains context and negotiates turns across multimodal exchanges so that state remains coherent across intermittent sessions. Conversation with LLM-based agents is not merely turn taking but it is the disciplined management of context over time. Robust dialogue practices such as greetings, confirmations, clarifications and summarization stabilize meaning and keep state coherent \cite{shaikh2023grounding}). Empirical studies of LLM voice assistants likewise show richer, more adaptive turn taking that mitigates breakdowns through proactive follow-ups \cite{mahmood2023llm}. In clinical and patient facing settings, such disciplined conduct reduces miscommunication, improves data capture and supports safer handoffs between encounters. Beyond direct chat, agent behavior must also be orchestrated by external signals like incoming lab results, device data streams or workflow events. In healthcare operations, such event-driven activation enables timely triage, order checks and monitoring escalations without relying on continuous user supervision. Because clinical decisions have consequences, interaction must explicitly allocate roles for oversight. Human-AI collaboration work in medicine shows that expert steerable tools improve trust and utility without sacrificing accuracy \cite{cai2019human}. Finally, interactions must anticipate failure and recover gracefully. Generic safeguards include greedy decoding such as self-consistency to expose divergent reasoning paths \cite{wang2022self}, Chain-of-Verification routines that plan and answer fact checking questions before issuing a final response \cite{dhuliawala2023chain} and reflective replanning that rewrites goals or tool sequences after unsuccessful trials \cite{shinn2023reflexion}. Operationally, resilient agents employ transactional rollbacks, bounded retries, circuit breakers and degraded safe modes. In clinical environments, such modality agnostic behaviors help avert propagation of errors such as halting downstream order placement after a failed formulary check, suppressing duplicate documentation after API timeouts etc. As compiled in Table \ref{tab:interaction}, Interaction Patterns are assessed along:
\label{interaction}

\normalsize
\subsubsection{Conversational Mode}

Conversational Mode characterizes an agent’s capacity for synchronous, turn-based exchanges that preserve discourse state across turns and sessions. Hallmarks include dialogue state persistence, session scoped memory for user intents and support for clarifying follow-ups rather than single-shot replies. Systems may expose a chat UI but are distinguished by continuity and resumability after interruptions. We rate this \cmark\ when the agent provides a turn-based conversational interface that maintains context across turns and supports resumptions. \dmark\ applies when multi-round interaction exists but is primarily a structured workflow or prompt sequenced exchange with fragile context retention. \xmark\ describes single query processing with no evidence of turn-based dialogue, session continuity or conversational context maintenance. In healthcare, virtual intake, symptom follow-ups and discharge education implemented systems reduce re-entry of information and support more efficient clinician-patient exchanges.

\normalsize
\subsubsection{Event-Triggered Activation}

Event-Triggered Activation denotes an agent’s capacity to autonomously monitor external signals via webhooks or message buses and initiate workflows when specified conditions are met without explicit user prompting. Mature designs subscribe to streams, apply temporal windows, enrich events with context and invoke idempotent handlers under rate limits. We rate this \cmark\ when real-time, automated triggers from external sources activate agents with continuous monitoring and proactive execution. \dmark\ is earned when basic triggering exists (API callbacks or scheduled polling) but activation is largely initiated by the user or lacks durable stream integration. \xmark\ is labeled when operation relies solely on prompted interactions or internal thresholds with no autonomous monitoring of external streams. For healthcare settings, robust activation delivers timely interventions like abnormal vitals, formulary aware order suppression etc.

\normalsize
\subsubsection{Human-in-the-Loop}

Human-in-the-Loop captures an agent’s explicit insertion of human control during execution and not merely at input or after output. This is achieved via confirmation gates, uncertainty-triggered pauses and rollback controls. The gathered feedback is then stored in an auditable manner to refine memory or downstream policies. We rate this \cmark\ when users can confirm, approve or redirect the agent mid-run. \dmark\ applies when only basic review or approval is simulated or confirmations are optional without durable feedback incorporation. \xmark\ describes pipelines with no in-process oversight. In workflows like prior-authorization and sensitive chart updates, robust Human-in-the-Loop limits unchecked autonomy, improves accountability and supports clinical governance.

\normalsize
\subsubsection{Error Recovery}

Error Recovery denotes an agent’s capacity to detect, contain and correct failures across parsing, planning, tool use and system integration. In this sub-dimension, the agent flags anomalies, conducts a structured diagnosis of likely causes and then applies corrective steps which can be issuing disambiguation prompts, switching to alternative tools or repairing parameters. All these corrective measures are taken ensuring transactional safety. Mature designs incorporate explicit recovery nodes that capture exceptions and route control back to an executor for re-execution. We rate this \cmark\ when robust detection, retries and fallback procedures are integrated as first class protocols with demonstrated re-execution paths. \dmark\ applies when only basic detection or limited fallbacks and retries are present. \xmark\ describes pipelines lacking explicit runtime error handling or recovery strategies. Robust recovery prevents error cascades in clinical care. It halts malformed order writes, suppresses duplicate documentation when APIs time out and pauses chart updates until verification succeeds.

\begin{table}
\caption{Evaluation of Interaction Patterns}
\setlength{\tabcolsep}{3pt}
\centering
\begin{tabular}{|p{60pt}|p{35pt}|p{40pt}|p{35pt}|p{32pt}|}

\hline

\textbf{Papers} & \textbf{Conver-sational Mode}  & \textbf{Event-Triggered Activation} & \textbf{Human-in-the Loop} & \textbf{Error Recovery}\\
\hline

Colacare \cite{wang2025colacare} & \xmark & \xmark & \xmark & \dmark\\
\hline
Medagent \cite{tang2023medagents} & \dmark & \xmark & \xmark & \dmark\\
\hline
Medaide \cite{yang2024medaide} & \cmark & \xmark & \xmark & \xmark \\
\hline
Doctoragent \cite{feng2025doctoragent} & \cmark & \xmark & \xmark & \xmark \\
\hline
Aihospital \cite{fan2024ai} & \cmark & \xmark & \xmark & \xmark \\
\hline
Mmedagent \cite{li2024mmedagent} & \cmark & \xmark & \xmark & \xmark \\
\hline
Medchat \cite{liu2025medchat} & \cmark & \xmark & \xmark & \xmark \\
\hline
Medco \cite{wei2024medco} & \dmark & \xmark & \xmark & \xmark \\
\hline
Medagentpro \cite{wang2025medagent} & \xmark & \xmark & \xmark & \dmark \\
\hline
Clinicalagent \cite{yue2024clinicalagent} & \cmark & \xmark & \xmark & \xmark \\
\hline
Ktas \cite{han2024development} & \xmark & \xmark & \xmark & \xmark\\
\hline
Agentmd \cite{jin2024agentmd} & \xmark & \xmark & \xmark & \dmark\\
\hline
Mdagent \cite{kim2024mdagents} & \cmark & \xmark & \xmark & \dmark\\
\hline
Agenthospital \cite{li2024agent} & \xmark & \cmark & \xmark & \dmark\\
\hline
Pharmagent \cite{gao2025pharmagents} & \xmark & \xmark & \xmark & \dmark\\
\hline
Txagent \cite{gao2025txagent} & \xmark & \xmark & \xmark & \dmark\\
\hline
Autoct \cite{liu2025autoct} & \xmark & \xmark & \xmark & \dmark\\
\hline
Kg4diagnosis \cite{zuo2025kg4diagnosis} & \xmark & \xmark & \cmark & \xmark\\
\hline
Mdteamgpt \cite{chen2025mdteamgpt} & \dmark & \xmark & \xmark & \dmark\\
\hline
Mitigating \cite{ke2024mitigating} & \cmark & \xmark & \xmark & \xmark\\
\hline
Polaris \cite{mukherjee2024polaris} & \cmark & \xmark & \cmark & \dmark\\
\hline
Rxstrategist \cite{van2024rx} & \xmark & \xmark & \xmark & \xmark\\
\hline
Clinicallab \cite{yan2024clinicallab} & \xmark & \xmark & \cmark & \xmark\\
\hline
Agenticllm \cite{sudarshan2024agentic} & \xmark  & \xmark  & \xmark  & \xmark \\
\hline
Surgbox \cite{wu2024surgbox} & \cmark & \cmark & \xmark & \xmark\\
\hline
Aipatient \cite{yu2024simulated} & \cmark & \xmark & \xmark & \dmark\\
\hline
Healthagent \cite{abbasian2023conversational} & \cmark & \xmark & \xmark & \xmark\\
\hline
Icdcoding \cite{li2024exploring} & \xmark & \xmark & \dmark & \xmark\\
\hline
Aha \cite{pandey2024advancing} & \xmark & \xmark & \xmark & \xmark\\
\hline
Ddo \cite{jia2025ddo} & \cmark & \xmark & \xmark & \xmark\\
\hline
Beyond \cite{wang2024beyond} & \xmark & \xmark & \xmark & \xmark\\
\hline
Zodiac \cite{zhou2024zodiac} & \xmark & \dmark & \xmark & \xmark\\
\hline
Fhirviz \cite{almutairi2024fhirviz} & \xmark & \xmark & \cmark & \cmark \\
\hline
Fhir \cite{de2024multi} & \xmark & \xmark & \xmark & \xmark \\
\hline
Fuasagent \cite{zhao2025autonomous} & \xmark & \xmark & \xmark & \xmark \\
\hline
Aitherapist \cite{wasenmuller2024script} & \cmark & \xmark & \xmark & \xmark \\
\hline
Radiology \cite{zeng2024enhancing} & \xmark & \xmark & \xmark & \dmark \\
\hline
Triageagent \cite{lu2024triageagent} & \cmark & \xmark & \xmark & \xmark \\
\hline
Piors \cite{bao2024piors} & \cmark & \xmark & \xmark & \xmark  \\
\hline
Selfevolve \cite{almansoori2025self} & \cmark & \xmark & \cmark & \dmark \\
\hline
Drugagent \cite{liu2024drugagent} & \xmark & \xmark & \xmark & \xmark\\
\hline
Ctagent \cite{mao2025ct} & \xmark & \xmark & \xmark & \xmark\\
\hline
Adagent \cite{hou2025adagent} & \cmark & \xmark & \xmark & \xmark\\
\hline
Drugpilot \cite{li2025drugpilot} & \cmark & \xmark & \cmark & \cmark\\
\hline
Llmadmin \cite{gebreab2024llm} & \cmark & \cmark & \xmark & \dmark\\
\hline
Swarm \cite{song2025llm} & \xmark & \xmark & \xmark & \dmark\\
\hline
Drugdiscover \cite{ock2025large} & \xmark & \xmark & \xmark & \xmark\\
\hline
Ragdrug \cite{lee2025rag} & \xmark & \xmark & \xmark & \xmark\\
\hline
Drugreason \cite{inoue2025drugagent} & \cmark & \xmark & \xmark & \dmark\\
\hline

\end{tabular}
\label{tab:interaction}
\end{table}

\normalsize
\subsection{Adaptation \& Learning}

Adaptation \& Learning is the discipline of keeping deployed agents calibrated to a moving world. It begins with continuous sensing for distributional and behavioral shifts, then applying targeted mitigations before clinical performance erodes. Drift can be detected by monitoring changes in text embedding distributions \cite{gupta2023measuring}, by activation delta probes \cite{abdelnabi2025get} and by black box statistical tests that audit behavior changes over time \cite{richter2024auditing}. Mitigations span prompt hotfixes, domain adaptation and scheduled retraining. Their necessity is evidenced by temporal dataset shifts degrading ICU prediction models in EHRs \cite{guo2022evaluation}. In practice, healthcare deployments couple these monitors with escalation and rollback to preserve patient safety and regulatory traceability. Beyond detection, agents must learn from evaluators. Reinforcement-based adaptation formalizes feedback as rewards. Research shows that constitutional self-supervision substitutes AI feedback for human labels to reduce harmfulness \cite{bai2022constitutional} and verbal reward agents store reflective critiques as episodic memories to steer subsequent decisions without weight updates \cite{shinn2023reflexion}. In clinical settings, these pathways enable clinician-in-the-loop reward shaping and post deployment alignment. For example, medical question answering systems refined with domain specific supervision and evaluation protocols achieve ground breaking reliability on USMLE style questions \cite{singhal2025toward}. Finally, rapid specialization under data scarcity relies on meta-learning and few-shot competence. Gradient-based meta-learners initialize models for fast adaptation from a handful of cases \cite{finn2017model}, while LLMs exhibit in-context learning that acquires new tasks from a few demonstrations at inference \cite{brown2020language}. Combined with retrieval and sparse labeling, these methods let agents generalize to a new hospital, imaging device or rare disease with minimal annotation burden. This is especially crucial where prospective data collection is slow and expensive. Table \ref{tab:adaptation} details Adaptation \& Learning results across its 3 sub-dimensions:

\begin{table}
\caption{Evaluation of Adaptation \& Learning}
\setlength{\tabcolsep}{3pt}
\centering
\begin{tabular}{|p{60pt}|p{45pt}|p{40pt}|p{40pt}|}

\hline

\textbf{Papers} & \textbf{Drift Detection \& Mitigation} & \textbf{Reinforce-ment Based Adaptation}   & \textbf{Meta-Learning \& Few-Shot}\\
\hline

Colacare \cite{wang2025colacare} & \xmark & \xmark & \cmark \\
\hline
Medagent \cite{tang2023medagents} & \xmark & \xmark & \cmark \\
\hline
Medaide \cite{yang2024medaide} & \xmark & \xmark & \xmark  \\
\hline
Doctoragent \cite{feng2025doctoragent} & \xmark & \cmark & \xmark  \\
\hline
Aihospital \cite{fan2024ai} & \xmark & \xmark & \xmark  \\
\hline
Mmedagent \cite{li2024mmedagent} & \xmark & \xmark & \xmark  \\
\hline
Medchat \cite{liu2025medchat} & \xmark & \xmark & \xmark  \\
\hline
Medco \cite{wei2024medco} & \xmark & \xmark & \xmark  \\
\hline
Medagentpro \cite{wang2025medagent} & \xmark & \xmark &\xmark  \\
\hline
Clinicalagent \cite{yue2024clinicalagent} & \xmark & \xmark & \cmark  \\
\hline
Ktas \cite{han2024development} & \xmark & \xmark &\xmark \\
\hline
Agentmd \cite{jin2024agentmd} & \xmark & \xmark &\xmark \\
\hline
Mdagent \cite{kim2024mdagents}  & \xmark & \xmark & \cmark  \\
\hline
Agenthospital \cite{li2024agent} & \xmark & \xmark &\xmark \\
\hline
Pharmagent \cite{gao2025pharmagents} & \xmark & \xmark &\xmark \\
\hline
Txagent \cite{gao2025txagent} & \xmark & \xmark &\xmark \\
\hline
Autoct \cite{liu2025autoct} & \xmark & \dmark &\xmark \\
\hline
Kg4diagnosis \cite{zuo2025kg4diagnosis} & \xmark & \cmark & \xmark \\
\hline
Mdteamgpt \cite{chen2025mdteamgpt} & \xmark & \xmark & \cmark \\
\hline
Mitigating \cite{ke2024mitigating} & \xmark & \xmark &\xmark \\
\hline
Polaris \cite{mukherjee2024polaris} & \dmark & \cmark & \xmark \\
\hline
Rxstrategist \cite{van2024rx} & \xmark & \xmark &\xmark \\
\hline
Clinicallab \cite{yan2024clinicallab} & \xmark & \xmark &\xmark \\
\hline
Agenticllm \cite{sudarshan2024agentic} & \xmark & \dmark & \xmark \\
\hline
Surgbox \cite{wu2024surgbox} & \xmark & \xmark &\xmark \\
\hline
Aipatient \cite{yu2024simulated} & \xmark & \xmark & \cmark \\
\hline
Healthagent \cite{abbasian2023conversational} & \xmark & \xmark &\xmark \\
\hline
Icdcoding \cite{li2024exploring} & \xmark & \xmark &\xmark \\
\hline
Aha \cite{pandey2024advancing} & \xmark & \xmark & \cmark \\
\hline
Ddo \cite{jia2025ddo} & \xmark & \cmark & \xmark \\
\hline
Beyond \cite{wang2024beyond} & \xmark  & \xmark  & \xmark  \\
\hline
Zodiac \cite{zhou2024zodiac} & \xmark & \xmark & \cmark \\
\hline
Fhirviz \cite{almutairi2024fhirviz} & \xmark  & \xmark  & \xmark  \\
\hline
Fhir \cite{de2024multi} & \xmark  & \xmark  & \xmark  \\
\hline
Fuasagent \cite{zhao2025autonomous} & \xmark  & \dmark  & \xmark \\
\hline
Aitherapist \cite{wasenmuller2024script} & \xmark  & \xmark  & \xmark  \\
\hline
Radiology \cite{zeng2024enhancing} &  \xmark  & \xmark  & \xmark  \\
\hline
Triageagent \cite{lu2024triageagent} & \xmark  & \xmark  & \xmark   \\
\hline
Piors \cite{bao2024piors} & \xmark  & \xmark  & \xmark   \\
\hline
Selfevolve \cite{almansoori2025self} & \xmark & \xmark & \cmark  \\
\hline
Drugagent \cite{liu2024drugagent} & \xmark  & \xmark  & \xmark \\
\hline
Ctagent \cite{mao2025ct} & \xmark  & \xmark  & \cmark \\
\hline
Adagent \cite{hou2025adagent} & \xmark  & \xmark  & \xmark\\
\hline
Drugpilot \cite{li2025drugpilot} & \xmark  & \xmark  & \xmark \\
\hline
Llmadmin \cite{gebreab2024llm} & \xmark  & \xmark  & \xmark \\
\hline
Swarm \cite{song2025llm} & \xmark  & \xmark  & \xmark  \\
\hline
Drugdiscover \cite{ock2025large} & \xmark & \cmark & \xmark \\
\hline
Ragdrug \cite{lee2025rag} & \xmark & \xmark & \dmark \\
\hline
Drugreason \cite{inoue2025drugagent} & \xmark & \xmark & \xmark \\
\hline

\end{tabular}
\label{tab:adaptation}
\end{table}

\normalsize
\subsubsection{Drift Detection \& Mitigation}

This sub-dimension concerns the capacity of LLM-based agents to remain reliable under evolving data distributions, task definitions or usage contexts. Drift manifests when the statistical profile of inputs diverges from the training regime or when task requirements subtly shift leading to performance degradation. In well implemented systems, statistical divergence tests on incoming data streams, embedding space comparisons or activation level drift classifiers provide early warnings. This is followed by solid mitigation through retraining, prompt adaptation or deployment of updated models constituting a \cmark\ state. In more limited cases, agents may exhibit only partial safeguards such as exposure to diverse samples during iterative training that reduces but does not directly monitor drift, depicting a \dmark\ state. Conversely, systems without explicit statistical monitoring or any corrective mechanisms remain vulnerable to silent degradation aligning with an \xmark\ state. In healthcare, where distributional shifts occur due to population demographics or evolving clinical guidelines, drift detection and mitigation is indispensable for preserving patient safety and minimizing diagnostic error.

\normalsize
\subsubsection{Reinforcement-Based Adaptation}

This sub-dimension captures how LLM-based agents refine their behavior by translating evaluative feedback into structured reward signals that shape future decisions. In fully fledged implementations, explicit reinforcement learning mechanisms such as RLHF are used. In these mechanisms, expert preferences are used to optimize policies that encode verbal feedback into memory for iterative improvement and make up a \cmark\ label. These systems demonstrate continuous adaptation where reward-driven loops refine decision policies and integrate expert validation. In contrast, partial implementations may rely on simplified forms of feedback such as binary success or failure signals, heuristic optimization, verbal reinforcement in prompts etc. These approaches provide directional guidance but lack systematic reinforcement mechanisms thereby portraying a \dmark\ label. Finally, systems that omit explicit reward modeling, reinforcement learning or preference-driven updates align with an \xmark\ category. In healthcare contexts, safe adaptation requires agents to integrate clinician feedback, regulatory requirements and patient specific preferences.

\normalsize
\subsubsection{Meta-Learning \& Few-Shot}

Meta-Learning \& Few-Shot refers to the ability of LLM-based agents to adapt quickly to new tasks or domains with minimal supervision. Meta-learning methods such as gradient-based algorithms that train models to update effectively from only a few examples enable rapid reconfiguration across tasks without requiring extensive retraining. Complementary to this, in-context few-shot prompting leverages pretrained representations to generalize at inference time where a handful of demonstrations within the prompt can guide accurate reasoning. Systems that combine these strategies using either meta-learned initialization or sophisticated few-shot prompting protocols are regarded as \cmark. More limited approaches such as relying solely on zero-shot generalization or retrieving exemplar cases without systematic adaptation are categorized as \dmark. Meanwhile, frameworks that show no mention of few-shot adaptation, meta-learning protocols or rapid task specialization are classified as \xmark. This sub-dimension's capabilities are particularly consequential in medicine where models must rapidly accommodate shifts in clinical practice, rare patient phenotypes and evolving evidence bases in dynamic care environments.


\normalsize
\subsection{Safety \& Ethics}

LLM agents in healthcare must proactively prevent unsafe behaviors through layered alignment, systematic red teaming and attack aware orchestration. Empirical work shows that jailbreaks can be auto-generated even when the model is a black box \cite{chao2025jailbreaking}. Recent studies document automatic prompt injection attacks \cite{liu2024automatic} and organize defenses into clear taxonomies \cite{yi2024jailbreak} while frameworks like Constitutional AI \cite{bai2022constitutional} and end-to-end red-teaming methods \cite{purpura2025building} establish procedural guardrails. Equity requires detecting and mitigating systematic errors across patient groups, with evaluation and remediation guided by LLM bias surveys \cite{gallegos2024bias,li2023survey}. In healthcare specifically, evidence such as the analysis by \cite{obermeyer2019dissecting} of a widely used risk algorithm illustrates how proxy targets can induce racial disparities, motivating stratified reporting and bias stress tests. Patient privacy hinges on minimizing identifiability and resisting model level leakage. Empirical attacks retrieve verbatim training data from models \cite{mireshghallah2022quantifying} underscoring the need for privacy-preserving learning. In healthcare, federated learning enables cross-institutional collaboration without centralizing PHI \cite{rieke2020future} complemented by robust clinical text deidentification. These practices enable model development, evaluation and monitoring while honoring confidentiality constraints at the point of care. Finally, trustworthy deployment demands verifiable compliance and auditability. Scholarly analyses detail how the EU AI Act reshapes risk classification and obligations for health AI \cite{van2024eu}. In the U.S., reviews of device oversight highlight gaps in transparency and representation for AI tools \cite{muralidharan2024scoping}. These frameworks ground provenance logging, rationale tracing and post market surveillance necessary for clinical accountability. Table \ref{tab:safety} presents Safety \& Ethics evaluations across its 4 sub-dimensions:

\begin{table}
\caption{Evaluation of Safety \& Ethics}
\setlength{\tabcolsep}{3pt}
\centering
\begin{tabular}{|p{60pt}|p{46pt}|p{29pt}|p{40pt}|p{40pt}|}

\hline

\textbf{Papers} & \textbf{Safety Guardrails \& Adversarial Robustness} & \textbf{Bias \& Fairness}   & \textbf{Privacy Preserving Mechanism} & \textbf{Regulatory \& Compliance Constraints}\\
\hline

Colacare \cite{wang2025colacare} & \xmark & \dmark & \dmark & \xmark\\
\hline
Medagent \cite{tang2023medagents} & \xmark & \xmark & \xmark & \xmark\\
\hline
Medaide \cite{yang2024medaide} & \dmark & \dmark & \dmark & \xmark \\
\hline
Doctoragent \cite{feng2025doctoragent} & \dmark & \xmark & \xmark & \cmark \\
\hline
Aihospital \cite{fan2024ai} & \xmark & \dmark & \dmark & \xmark \\
\hline
Mmedagent \cite{li2024mmedagent} & \xmark & \xmark & \xmark & \xmark \\
\hline
Medchat \cite{liu2025medchat} & \xmark & \xmark & \xmark & \xmark \\
\hline
Medco \cite{wei2024medco} & \xmark & \xmark & \dmark & \xmark \\
\hline
Medagentpro \cite{wang2025medagent} & \xmark & \xmark & \xmark & \xmark \\
\hline
Clinicalagent \cite{yue2024clinicalagent} & \xmark & \xmark & \xmark & \xmark \\
\hline
Ktas \cite{han2024development} & \xmark & \xmark & \cmark & \cmark \\
\hline
Agentmd \cite{jin2024agentmd} & \xmark & \xmark & \dmark & \dmark\\
\hline
Mdagent \cite{kim2024mdagents} & \xmark & \dmark & \xmark & \xmark\\
\hline
Agenthospital \cite{li2024agent} & \xmark & \dmark & \xmark & \xmark\\
\hline
Pharmagent \cite{gao2025pharmagents} & \xmark & \xmark & \xmark & \xmark \\
\hline
Txagent \cite{gao2025txagent} & \dmark & \dmark & \cmark & \xmark \\
\hline
Autoct \cite{liu2025autoct} & \xmark & \xmark & \xmark & \xmark \\
\hline
Kg4diagnosis \cite{zuo2025kg4diagnosis} & \dmark & \xmark & \xmark & \xmark\\
\hline
Mdteamgpt \cite{chen2025mdteamgpt} & \cmark & \dmark & \xmark & \xmark\\
\hline
Mitigating \cite{ke2024mitigating} & \xmark & \cmark & \xmark & \xmark\\
\hline
Polaris \cite{mukherjee2024polaris} & \cmark & \cmark & \cmark & \xmark\\
\hline
Rxstrategist \cite{van2024rx} & \xmark & \xmark & \dmark & \xmark\\
\hline
Clinicallab \cite{yan2024clinicallab} & \dmark & \cmark & \dmark & \dmark\\
\hline
Agenticllm \cite{sudarshan2024agentic} & \dmark & \xmark & \xmark & \xmark\\
\hline
Surgbox \cite{wu2024surgbox} & \dmark & \dmark & \dmark & \xmark\\
\hline
Aipatient \cite{yu2024simulated} & \dmark & \dmark & \cmark & \cmark\\
\hline
Healthagent \cite{abbasian2023conversational} & \xmark & \dmark & \dmark & \xmark\\
\hline
Icdcoding \cite{li2024exploring} & \xmark & \xmark & \dmark & \xmark\\
\hline
Aha \cite{pandey2024advancing} & \dmark & \xmark & \dmark & \xmark\\
\hline
Ddo \cite{jia2025ddo} & \xmark & \xmark & \xmark & \xmark\\
\hline
Beyond \cite{wang2024beyond} & \xmark & \xmark & \cmark & \xmark\\
\hline
Zodiac \cite{zhou2024zodiac} & \dmark & \cmark & \cmark & \cmark\\
\hline
Fhirviz \cite{almutairi2024fhirviz} & \dmark & \xmark & \xmark & \xmark \\
\hline
Fhir \cite{de2024multi} & \xmark & \xmark & \cmark & \xmark \\
\hline
Fuasagent \cite{zhao2025autonomous} & \dmark & \dmark & \cmark & \cmark \\
\hline
Aitherapist \cite{wasenmuller2024script} & \xmark & \xmark & \xmark & \xmark \\
\hline
Radiology \cite{zeng2024enhancing} & \dmark & \xmark & \xmark & \xmark \\
\hline
Triageagent \cite{lu2024triageagent} & \xmark & \xmark & \dmark & \xmark \\
\hline
Piors \cite{bao2024piors} & \xmark & \dmark & \dmark & \xmark \\
\hline
Selfevolve \cite{almansoori2025self} & \xmark & \cmark & \xmark & \xmark \\
\hline
Drugagent \cite{liu2024drugagent} & \xmark & \xmark & \xmark & \xmark\\
\hline
Ctagent \cite{mao2025ct} & \xmark & \xmark & \xmark & \xmark\\
\hline
Adagent \cite{hou2025adagent} & \xmark & \xmark & \cmark & \xmark\\
\hline
Drugpilot \cite{li2025drugpilot} & \dmark & \xmark & \xmark  & \xmark\\
\hline
Llmadmin \cite{gebreab2024llm} & \cmark & \xmark & \dmark & \xmark\\
\hline
Swarm \cite{song2025llm} & \xmark & \xmark & \xmark & \xmark\\
\hline
Drugdiscover \cite{ock2025large} & \xmark & \xmark & \xmark & \xmark\\
\hline
Ragdrug \cite{lee2025rag} & \xmark & \xmark & \xmark & \xmark\\
\hline
Drugreason \cite{inoue2025drugagent} & \xmark & \xmark & \xmark & \xmark\\
\hline

\end{tabular}
\label{tab:safety}
\end{table}

\normalsize
\subsubsection{Safety Guardrails \& Adversarial Robustness}

Safety Guardrails \& Adversarial Robustness defines the defense controls that prevent, detect and contain harmful behavior when agents are steered intentionally or accidentally away from intended clinical use. It spans hardening of inputs to prevent anything like prompt injection, resilient planning and output governance with continuous stress testing. Under this sub-dimension, \cmark\ denotes mature, multi-stage safeguards with automated detectors, adversarial evaluation and rapid remediation pathways. \dmark\ indicates isolated filters or ad hoc human review without systematic testing. \xmark\ reflects the absence of explicit mechanisms or evidence of effectiveness. In healthcare delivery, the distinction between these labels is practical. Robust systems block unsafe orders, resist manipulation of EHR linked tools, catch dosing anomalies and prevent escalation of failures across multi-step workflows.

\normalsize
\subsubsection{Bias \& Fairness}

Bias \& Fairness concerns whether an agent’s behavior remains equitable across clinically salient populations and contexts. It requires: (i) representative data with verified labels (ii) modeling choices that constrain disparate error (iii) deployment policies to minimize inequity. Within this sub-dimension, \cmark\ denotes balanced representation across clinically salient cohorts like age and gender. It also comprises of subgroup bias analyses with reported disparities and applied mitigation. \dmark\ applies when only privacy steps such as deidentification/anonymization are used or bias concerns are merely acknowledged. But there is no demographic auditing, no fairness metrics and no implemented bias detection or mitigation mechanisms. \xmark\ describes the absence of bias assessment and fairness mechanisms completely. In healthcare, bias and fairness is essential to prevent underdiagnosis or inequitable access particularly across intersections of race, age, sex, language, disability and socioeconomic status.

\normalsize
\subsubsection{Privacy-Preserving Mechanism}

This sub-dimension specifies the architectural and procedural controls that restrict who and what can access patient data across the agent lifecycle. This sub-dimension broadly covers mechanisms including edge inference, trusted execution, encryption in transit, private retrieval with scoped context windows, deidentification/pseudonymization and differential privacy for training as well as analytics. Programs should also test privacy explicitly via threat modeling, leakage audits and redaction verification. \cmark\ indicates agent level privacy mechanisms beyond dataset preprocessing entailing on-device handling of sensitive inputs, built in deidentification or anonymization, strong encryption and IRB/ethics approval when human or clinical data are involved. Many implementations also use a dual layer design that keeps sensitive processing local. \dmark\ applies when privacy is limited to dataset deidentification or general compliance claims (e.g., HIPAA compliant). This does not incorporate an agent side deidentification/anonymization mechanism  or technical controls such as encryption, differential privacy and secure computation. \xmark\ describes systems with no technical privacy controls as defined above. In clinical deployments, these controls prevent PHI exfiltration from EHR linked tools, reduce insider exposure and enable compliant collaboration without centralizing sensitive data.

\normalsize
\subsubsection{Regulatory \& Compliance Constraints}

Regulatory \& Compliance Constraints specifies how an agent’s design, data handling and operating procedures satisfy binding laws across collection, training, inference and post deployment monitoring. Core elements include: (i) documented consent and lawful bases (ii) role scoped access and retention schedules (iii) cross-border transfer controls (iv) formal risk assessments that tie policies to technical safeguards. Compliance should be evidenced and not merely asserted through policies, test records and release gates. In healthcare, these controls determine whether an agent’s recommendation features would be treated as regulated software, whether PHI handling meets HIPAA or GDPR duties and whether deployments withstand clinical legal review. \cmark\ indicates clear alignment with applicable laws like HIPAA or GDPR, operational consent and retention controls. It also encompasses IRB or ethics approval when human subjects or clinical data are involved and incident response procedures that include documented drills. \dmark\ applies when evidence is limited to isolated attestations like HIPAA compliant cloud and third party certifications. This does not cover documented consent flows or risk assessment with policy statements that lack operational evidence and reliance on vendor assurances. \xmark\ describes systems with no explicit mapping to laws or standards.


\normalsize
\subsection{Framework Typology}

Framework Typology delineates the structural and operational blueprints underlying LLM-based agent systems. It articulates how multiple reasoning units are composed and how control is exercised over their joint behavior. In many recent systems, we see a design that partitions cognitive labor across several specialized agent roles which collaborate via rounds of discussion to reach consensus. For example, \cite{tang2023medagents} uses a multi-round discussion among LLM agents that mimic domain experts, combining their individual analyses into a summary before arriving at a decision. Similarly, the Multi-Agent Conversation (MAC) \cite{chen2025enhancing} framework emulates multidisciplinary team discussions in rare disease diagnosis. It involves multiple doctor agents plus a supervisor agent discussing clinical features and different diagnostic hypotheses. Systems of this kind yield benefits in medical question answering, clinical decision support and diagnosis because they mirror multidisciplinary workflows. They contrast with monolithic designs where a single LLM is responsible for all stages of reasoning which may be simpler but risk domain misknowledge in healthcare settings. Complementing the composition of agent roles is the mechanism by which control, state, feedback and adaptation are governed over time. This includes how tasks are orchestrated and whether there is a central controller that assigns subtasks, monitors agent outputs and enforces checks. For instance, \cite{zhang2025agentorchestra} implements a central planning agent that breaks down complex objectives and delegates to sub-agents, then aggregates and synthesizes their outputs. Such centralized orchestration is particularly crucial in clinical environments to ensure compliance, to manage latency and to integrate human interventions at critical junctures. On the other hand, some frameworks explore decentralized or adaptive orchestration. Along those lines, \cite{yang2025agentnet} lets agents self-specialize, evolve their connections in a DAG structure and route tasks without a single central controller. Table \ref{tab:framework} provides Framework Typology assessments across:

\normalsize
\subsubsection{Multi-Agent Design}

Multi-Agent Design captures how responsibilities within an LLM-based system are distributed across one or multiple agents and how these agents interact to accomplish tasks. At one extreme, single-agent designs encapsulate reasoning, planning and execution within a unified process providing us with simplicity but limiting specialization. More advanced systems decompose functionality into distinct agents such as planners, retrievers or verifiers. These distinct agents then exchange messages or deliberate collectively, thereby enabling modularity, redundancy and clearer alignment with domain specific expertise. In healthcare, such distributed structures can emulate multidisciplinary collaboration. For instance, having separate agents to parse imaging results, validate treatment guidelines and draft patient facing summaries resembles such multidisciplinary collaboration. Evaluating this sub-dimension requires distinguishing whether a system demonstrates a genuine multi-agent architecture, only partial traces of such coordination or no such implementation at all. Thus, \cmark\ applies when specialized roles are clearly instantiated, \dmark\ captures ambiguous or superficial coordination and \xmark\ denotes monolithic or tool-augmented single-agent systems.

\normalsize
\subsubsection{Centralized Orchestration}

Centralized Orchestration refers to how control is exercised over the flow of tasks, data and decisions in LLM-based agent systems. In its strongest form, a central controller manages the sequencing of agents, monitors their outputs and integrates results into a coherent whole. Although, centralization can introduce single points of failure and latency bottlenecks, it ensures global consistency, enables oversight and provides clear points of intervention. Other frameworks rely on looser coordination such as hierarchical delegation or peer-to-peer communication where agents collaborate without a single coordinating authority. These offer flexibility but may reduce reliability in safety critical settings. Some systems operate with only sequential task execution and minimal oversight, resembling simple pipelines rather than true orchestration. In healthcare, these distinctions carry significant weight. A robust orchestrator can ensure that multiple evidence sources are reconciled before clinical decisions are proposed and trigger escalation to human experts when uncertainty is high. Without such orchestration, systems risk fragmented reasoning, missed contradictions and limited accountability. Operationally, orchestration is often realized via workflow engines and policy-based schedulers that implement timeouts, retries and circuit breakers. The evaluation of this sub-dimension rests on the sophistication of coordination. \cmark\ applies when a central orchestrator is explicitly defined and bi-directional control is visible. \dmark\ captures looser structures such as peer-to-peer or hierarchical delegation. \xmark\ describes sequential or single-agent designs.

\begin{table}
\caption{Evaluation of Framework Typology}
\setlength{\tabcolsep}{3pt}
\centering
\begin{tabular}{|p{60pt}|p{45pt}|p{50pt}|}

\hline

\textbf{Papers} & \textbf{Multi-Agent Design} & \textbf{Centralized Orchestration}\\
\hline

Colacare \cite{wang2025colacare} & \cmark & \cmark  \\
\hline
Medagent \cite{tang2023medagents} & \cmark & \dmark  \\
\hline
Medaide \cite{yang2024medaide} & \cmark & \cmark   \\
\hline
Doctoragent \cite{feng2025doctoragent} & \cmark & \cmark   \\
\hline
Aihospital \cite{fan2024ai} & \cmark & \dmark   \\
\hline
Mmedagent \cite{li2024mmedagent} & \xmark & \xmark \\
\hline
Medchat \cite{liu2025medchat} & \cmark & \dmark   \\
\hline
Medco \cite{wei2024medco} & \cmark & \dmark  \\
\hline
Medagentpro \cite{wang2025medagent} & \xmark & \dmark   \\
\hline
Clinicalagent \cite{yue2024clinicalagent} & \cmark & \cmark   \\
\hline
Ktas \cite{han2024development} & \cmark & \dmark \\
\hline
Agentmd \cite{jin2024agentmd} & \xmark & \xmark \\
\hline
Mdagent \cite{kim2024mdagents} & \cmark & \cmark \\
\hline
Agenthospital \cite{li2024agent} & \cmark & \cmark \\
\hline 
Pharmagent \cite{gao2025pharmagents} & \cmark & \dmark \\
\hline
Txagent \cite{gao2025txagent} & \cmark & \dmark \\
\hline
Autoct \cite{liu2025autoct} & \cmark & \dmark  \\
\hline
Kg4diagnosis \cite{zuo2025kg4diagnosis} & \cmark & \cmark \\
\hline
Mdteamgpt \cite{chen2025mdteamgpt} & \cmark & \cmark \\
\hline
Mitigating \cite{ke2024mitigating} & \cmark & \dmark \\
\hline
Polaris \cite{mukherjee2024polaris} & \cmark & \cmark \\
\hline
Rxstrategist \cite{van2024rx} & \cmark & \dmark \\
\hline
Clinicallab \cite{yan2024clinicallab} & \cmark & \dmark  \\
\hline
Agenticllm \cite{sudarshan2024agentic} & \dmark & \xmark \\
\hline
Surgbox \cite{wu2024surgbox} & \cmark & \cmark \\
\hline
Aipatient \cite{yu2024simulated} & \cmark & \dmark \\
\hline
Healthagent \cite{abbasian2023conversational} & \xmark & \xmark \\
\hline
Icdcoding \cite{li2024exploring} & \cmark & \dmark \\
\hline
Aha \cite{pandey2024advancing} & \cmark & \dmark \\
\hline
Ddo \cite{jia2025ddo} & \cmark & \dmark \\
\hline
Beyond \cite{wang2024beyond} & \cmark & \dmark \\
\hline
Zodiac \cite{zhou2024zodiac} & \cmark & \dmark \\
\hline
Fhirviz \cite{almutairi2024fhirviz} & \cmark & \dmark\\
\hline
Fhir \cite{de2024multi} & \cmark & \cmark \\
\hline
Fuasagent \cite{zhao2025autonomous} & \cmark & \dmark \\
\hline
Aitherapist \cite{wasenmuller2024script} & \cmark & \dmark \\
\hline
Radiology \cite{zeng2024enhancing} & \cmark & \dmark \\
\hline
Triageagent \cite{lu2024triageagent} & \cmark & \dmark \\
\hline
Piors \cite{bao2024piors} & \cmark & \dmark \\
\hline
Selfevolve \cite{almansoori2025self} & \cmark & \dmark \\
\hline
Drugagent \cite{liu2024drugagent} & \cmark & \dmark \\
\hline
Ctagent \cite{mao2025ct} & \xmark & \xmark \\
\hline
Adagent \cite{hou2025adagent} & \xmark & \xmark \\
\hline
Drugpilot \cite{li2025drugpilot} & \xmark & \xmark \\
\hline
Llmadmin \cite{gebreab2024llm} & \cmark & \cmark \\
\hline
Swarm \cite{song2025llm} & \cmark & \dmark \\
\hline
Drugdiscover \cite{ock2025large} & \xmark & \xmark \\
\hline
Ragdrug \cite{lee2025rag} & \cmark & \dmark \\
\hline
Drugreason \cite{inoue2025drugagent} & \cmark & \cmark \\
\hline
\end{tabular}
\label{tab:framework}
\end{table}

\normalsize
\subsection{Core Tasks \& Subtasks}

Core tasks and subtasks define what an LLM-based agent actually does in healthcare. They scope the agent’s operational responsibilities and the verifiable outputs it must deliver. Starting from raw clinical data, these tasks progress toward clinical recommendations and system level evaluation. This dimension therefore grounds agent design in concrete workflows and measurable endpoints. To start with, agents can transform unstructured records into computable artifacts. They can summarize problems, extract entities and link to codes. They can reason over multi-table EHRs to answer record level queries as shown by EHRAgent \cite{shi2024ehragent}. Multi-agent pipelines can mirror documentation roles as well. For example, ColaCare \cite{wang2025colacare} coordinates DoctorAgents with a supervising MetaAgent to integrate structured EHR data and text reasoning. \cite{pandey2024advancing} frames prior-authorization evidence extraction as a multi-agent classification and retrieval problem. These designs connect directly to everyday charting, coding and utilization management. Next, agents can answer clinical questions and broker guideline-based support. MDAgents \cite{kim2024mdagents} adapts agent roles moderator, retriever, verifier to task complexity and shows gains across medical decision benchmarks. For EHR-grounded questions, agents plan tool use and compute evidence inside the chart as in EHRAgent \cite{shi2024ehragent}. This pattern ties answers to provenance which clinicians need for accountable decision support. Further, agents can conduct triage and seed differentials under uncertainty. Role specialized teams emulate emergency clinicians, pharmacists and coordinators. \cite{han2024development} shows how a multi-agent clinical decision support system aligned to the Korean Triage and Acuity Scale operationalizes triage, treatment planning and routing in the emergency department. Hospital simulations let agents practice intake, inquiry and differential ranking end-to-end as in \cite{li2024agent}. These approaches reflect the time pressure, missingness and bias checks typical of frontline care. Next, agents can perform diagnostic reasoning through plans, tool calls and explicit updates to hypotheses. Hierarchical teams split generalist intake from subspecialist review while grounding steps in knowledge graphs as discussed in \cite{zuo2025kg4diagnosis}. MedAgent-Pro \cite{wang2025medagent} showcases a multimodal reasoning agent for diagnosis. Modular designs target interactive differential diagnosis with an orchestrator and role agents as the authors of MEDDxAgent show in \cite{rose2025meddxagent}. This mirrors consult pathways where uncertainty tracking and escalation are mandatory. Continuing, agents can plan treatments and verify prescriptions. Frameworks focused on therapeutics treat tool use as a first class operation. TxAgent \cite{gao2025txagent} selects from a ToolUniverse to check contraindications and personalize options. Prescription verification can be staged across indication, dose and interaction checks with an agentic pipeline as documented in Rx Strategist \cite{van2024rx}. Conversational reconciliation agents bring patients into the loop and reconcile lists before orders are signed \cite{deo2025conversational}. These steps align with order sets, pharmacy verification and stewardship policies in practice. Agents can also accelerate discovery and clinical trials. In discovery, chemistry agents plan syntheses and reason over lab tools as illustrated by ChemCrow \cite{m2024augmenting}. In trials, multi-agent systems support outcome prediction and study analytics \cite{yue2024clinicalagent}. Patient-trial matching benefits from agentic knowledge augmentation \cite{shi2024enhancing} and complementary retrieval + LLM approaches such as TrialGPT \cite{jin2024matching}. These functions shorten candidate screening and increase equitable access to studies. Finally, agents can sustain patient interaction and monitoring between visits. Wearable systems compute personalized insights via agentic code execution and retrieval as the experiments show in \cite{merrill2024transforming}. Clinical deployments study physician supervised agents for medication reconciliation and safe communication \cite{deo2025conversational}. These designs map to nurse line triage, remote monitoring programs and digital chronic care pathways. Lastly, agents can be benchmarked and stress tested in interactive, tool rich environments. Simulation testbeds evaluate end-to-end decisions, dialogue and tool use with realistic artifacts. AgentClinic \cite{schmidgall2024agentclinic} measures agent behavior in multimodal clinical encounters across specialties and languages. MedAgentBench \cite{jiang2025medagentbench} embeds agents in a FHIR compliant virtual EHR with hundreds of physician written tasks. Such platforms link agent behavior to safety, external validity and workflow impact the way health systems evaluate new tools. Tables \ref{tab:task1} and Table \ref{tab:task2} present the Core Tasks \& Subtasks evaluations across:

\normalsize
\subsubsection{Clinical Documentation \& EHR Analysis}
This sub-dimension evaluates an agent’s ability to convert heterogeneous clinical documentation and EHR artifacts into computable and clinician usable outputs. Characteristics of this sub-dimension span chart summarization, extraction of clinically salient entities, mapping to codes like ICD, risk prediction from longitudinal signals and draft generation of reports like radiology impressions. An agent merits \cmark\ when it demonstrates clear medical record processing and structured information extraction on real clinical data and not just demos. This could include supporting core EHR analysis, clinical prediction tasks, reading from and writing to FHIR and handling specialty reports. \dmark\ indicates basic record generation or narrow document handling like prescriptions only and/or reliance on simulated rather than real documentation. \xmark\ applies when no EHR analysis is present.

\normalsize
\subsubsection{Medical Question Answering \& Decision Support}

Medical Question Answering \& Decision Support determines whether an agent can answer clinical questions and deliver guideline-grounded decision support across modalities and workflows. Capabilities include: (i) factoid and synthesis question answering over authoritative medical sources (ii) retrieval-augmented recommendations (iii) multi-turn diagnostic questioning (iv) task specific support such as triage classification, cardiology assessment, surgical planning advice and visual question answering. \cmark\ applies when either capability medical question answering or decision support is clearly realized. This could include domain specific scientific question answering with benchmark validation, evidence-based answers tied to citations, structured diagnostic processes, real-time recommendations, visual question answering for radiology regions and systematic reasoning traces. \dmark\ captures narrow or limited scope especially when the demonstrations are simulation-based only. Prior-authorization medical necessity checks, prescription verification questions, dialogue without robust guideline retrieval are good examples for \dmark. \xmark\ denotes the absence of medical question answering or decision support.

\normalsize
\subsubsection{Triage \& Differential Diagnosis}

This sub-dimension examines an agent’s capability to turn first contact information into two outputs: an acuity decision and a defensible list of alternatives. The agent elicits the chief complaint and history. It synthesizes vitals along with context and asks targeted follow-ups. It aligns to established triage scales like ESI, KTAS and recommends a receiving service or department with a brief rationale. In parallel, it constructs a differential diagnosis, lists key discriminators and proposes next tests. \cmark\ applies when either triage or differential diagnosis is clearly implemented. This includes symptom processing with ESI/KTAS-based urgency, department recommendations with reasoning, production of ranked alternatives and safe routing via role agents. \dmark\ captures narrow diagnosis without explicit prioritization, single condition classifiers, probability lists without urgency or simulation only workflows. \xmark\ denotes the absence of these functions. In clinical care, these capabilities directly govern disposition, time to treatment and escalation.

\normalsize
\subsubsection{Diagnostic Reasoning}

Diagnostic Reasoning tests an agent’s capacity to plan and execute multi-step clinical inference that links targeted evidence gathering to a defensible final diagnosis. Core behaviors include: (i) eliciting and structuring new data (ii) selecting and sequencing diagnostic tests (iii) interpreting results from structured and unstructured sources (iv) maintaining and revising hypotheses (v) fusing inputs from specialist agents (vi) aligning conclusions with guidelines (vii) traversing knowledge graphs (viii) calibrating confidence with explicit discriminators. An agent merits \cmark\ when it exhibits a structured reasoning workflow, performs active data acquisition, ensembles specialist opinions via principled decision fusion and outputs ranked differentials with rationale. \dmark\ captures shallow or single-shot classification, limited aggregation or region specific heuristics. \xmark\ denotes no diagnostic inference beyond extraction.

\begin{table}
\caption{Evaluation of Core Tasks \& Subtasks Part 1}
\setlength{\tabcolsep}{3pt}
\centering
\begin{tabular}{|p{60pt}|p{35pt}|p{40pt}|p{40pt}|p{36pt}|}

\hline

    \textbf{Papers} & \textbf{Clinical Documentation \& EHR Analysis} & \textbf{Medical Question Answering \& Decision Support}  & \textbf{Triage \& Differential Diagnosis} & \textbf{Diagnostic Reasoning} \\
\hline

Colacare \cite{wang2025colacare} & \cmark & \cmark & \dmark & \dmark\\
\hline
Medagent \cite{tang2023medagents} & \cmark & \cmark & \dmark & \dmark\\
\hline
Medaide \cite{yang2024medaide} & \dmark & \cmark & \cmark & \cmark \\
\hline
Doctoragent \cite{feng2025doctoragent} & \xmark & \cmark & \cmark & \cmark \\
\hline
Aihospital \cite{fan2024ai} & \cmark & \dmark & \dmark & \cmark \\
\hline
Mmedagent \cite{li2024mmedagent} & \cmark & \cmark & \xmark & \dmark \\
\hline
Medchat \cite{liu2025medchat} & \dmark & \dmark & \dmark & \cmark \\
\hline
Medco \cite{wei2024medco} & \dmark & \dmark & \dmark & \cmark \\
\hline
Medagentpro \cite{wang2025medagent} & \xmark & \cmark & \cmark & \cmark \\
\hline
Clinicalagent \cite{yue2024clinicalagent} & \xmark & \xmark & \xmark & \xmark \\
\hline
Ktas \cite{han2024development} & \xmark & \cmark & \cmark & \dmark\\
\hline
Agentmd \cite{jin2024agentmd} & \cmark & \cmark & \dmark & \dmark\\
\hline
Mdagent \cite{kim2024mdagents} & \cmark & \cmark & \cmark & \cmark\\
\hline
Agenthospital \cite{li2024agent} & \dmark & \cmark & \cmark & \cmark\\
\hline
Pharmagent \cite{gao2025pharmagents} & \xmark & \xmark & \xmark & \xmark \\
\hline
Txagent \cite{gao2025txagent} & \xmark & \cmark & \xmark & \xmark \\
\hline
Autoct \cite{liu2025autoct} & \xmark & \xmark & \xmark & \xmark\\
\hline
Kg4diagnosis \cite{zuo2025kg4diagnosis} & \dmark & \cmark & \cmark & \cmark\\
\hline
Mdteamgpt \cite{chen2025mdteamgpt} & \xmark & \cmark & \cmark & \cmark\\
\hline
Mitigating \cite{ke2024mitigating} & \dmark & \cmark & \cmark & \cmark\\
\hline
Polaris \cite{mukherjee2024polaris} & \cmark & \cmark & \cmark & \cmark\\
\hline
Rxstrategist \cite{van2024rx} & \dmark & \dmark & \xmark & \dmark\\
\hline
Clinicallab \cite{yan2024clinicallab} & \cmark & \cmark & \cmark & \cmark\\
\hline
Agenticllm \cite{sudarshan2024agentic} & \cmark & \xmark & \xmark & \xmark\\
\hline
Surgbox \cite{wu2024surgbox} & \dmark & \cmark & \xmark & \xmark\\
\hline
Aipatient \cite{yu2024simulated} & \cmark & \cmark & \xmark & \xmark\\
\hline
Healthagent \cite{abbasian2023conversational} & \cmark & \cmark & \xmark & \xmark\\
\hline
Icdcoding \cite{li2024exploring} & \cmark & \xmark & \xmark & \dmark\\
\hline
Aha \cite{pandey2024advancing} & \cmark & \dmark & \xmark & \dmark\\
\hline
Ddo \cite{jia2025ddo} & \cmark & \dmark & \cmark & \cmark\\
\hline
Beyond \cite{wang2024beyond} & \xmark & \cmark & \dmark & \cmark\\
\hline
Zodiac \cite{zhou2024zodiac} & \cmark & \cmark & \dmark & \cmark\\
\hline
Fhirviz \cite{almutairi2024fhirviz} & \cmark & \xmark & \xmark & \xmark \\
\hline
Fhir \cite{de2024multi} & \cmark & \xmark & \xmark & \xmark \\
\hline
Fuasagent \cite{zhao2025autonomous} & \dmark & \dmark & \xmark & \dmark \\
\hline
Aitherapist \cite{wasenmuller2024script} & \xmark & \dmark & \xmark & \xmark \\
\hline
Radiology \cite{zeng2024enhancing} & \cmark & \xmark & \xmark & \cmark \\
\hline
Triageagent \cite{lu2024triageagent} & \cmark & \cmark & \cmark & \dmark \\
\hline
Piors \cite{bao2024piors} & \cmark & \cmark & \cmark & \dmark \\
\hline
Selfevolve \cite{almansoori2025self} & \dmark & \cmark & \dmark & \cmark \\
\hline
Drugagent \cite{liu2024drugagent} & \xmark & \xmark & \xmark & \xmark\\
\hline
Ctagent \cite{mao2025ct} & \cmark & \cmark & \dmark & \dmark\\
\hline
Adagent \cite{hou2025adagent} & \dmark & \cmark & \dmark & \dmark\\
\hline
Drugpilot \cite{li2025drugpilot} & \xmark & \xmark & \xmark & \xmark\\
\hline
Llmadmin \cite{gebreab2024llm} & \cmark & \xmark & \xmark & \xmark\\
\hline
Swarm \cite{song2025llm} & \xmark & \xmark & \xmark & \xmark\\
\hline
Drugdiscover \cite{ock2025large} & \xmark & \cmark & \xmark & \xmark\\
\hline
Ragdrug \cite{lee2025rag} & \dmark & \cmark & \xmark & \xmark\\
\hline
Drugreason \cite{inoue2025drugagent} & \xmark & \xmark & \xmark & \xmark\\
\hline
    
\end{tabular}
\label{tab:task1}
\end{table}

\normalsize
\subsubsection{Treatment Planning \& Prescription}

Treatment Planning \& Prescription concerns an agent’s ability to translate diagnostic context into an individualized treatment plan and executable orders. Functionalities include selecting pharmacologic and non-pharmacologic options, personalizing dose, checking contraindications, reconciling active medications and generating patient counseling with stop criteria. Robust systems also account for allergies, renal function, comorbidities and formularies. They can also provide shared decision prompts, record provenance and hand off to a pharmacist or prescriber for verification. \cmark\ is warranted when either treatment planning or prescription is clearly implemented. Under this label, the system provides explicit recommendations, interaction checks, agentic pharmacist review or tool-based safety checks and rationale tied to patient factors. \dmark\ signals basic option selection or templated suggestions without dose optimization, interaction screening or personalization. It can also have demonstrations limited to simulation or education with mandatory human interpretation. \xmark\ is assigned in cases with no treatment recommendation or prescription management.

\begin{table}
\caption{Evaluation of Core Tasks \& Subtasks Part 2}
\setlength{\tabcolsep}{3pt}
\centering
\begin{tabular}{|p{60pt}|p{37pt}|p{39pt}|p{35pt}|p{39pt}|}

\hline
    \textbf{Papers} & \textbf{Treatment Planning \& Prescription} & \textbf{Drug Discovery \& Clinical Trial Design}  & \textbf{Patient Interaction \& Monitoring} & \textbf{Bench-marking \& Simulation Environment} \\
    
\hline

Colacare \cite{wang2025colacare} & \xmark & \xmark & \xmark & \cmark\\
\hline
Medagent \cite{tang2023medagents} & \xmark & \xmark & \xmark & \cmark\\
\hline
Medaide \cite{yang2024medaide} & \cmark & \xmark & \dmark & \cmark \\
\hline
Doctoragent \cite{feng2025doctoragent} & \cmark & \xmark & \cmark & \cmark \\
\hline
Aihospital \cite{fan2024ai} & \cmark & \xmark & \cmark & \cmark \\
\hline
Mmedagent \cite{li2024mmedagent} & \xmark & \xmark & \xmark & \cmark \\
\hline
Medchat \cite{liu2025medchat} & \dmark & \xmark & \cmark & \xmark \\
\hline
Medco \cite{wei2024medco} & \dmark & \xmark & \dmark & \cmark \\
\hline
Medagentpro \cite{wang2025medagent} & \xmark & \xmark & \xmark & \cmark \\
\hline
Clinicalagent \cite{yue2024clinicalagent} & \xmark & \cmark & \xmark & \dmark \\
\hline
Ktas \cite{han2024development} & \cmark & \xmark & \xmark & \cmark\\
\hline
Agentmd \cite{jin2024agentmd} & \xmark & \xmark & \xmark & \cmark\\
\hline
Mdagent \cite{kim2024mdagents} & \xmark & \xmark & \xmark & \cmark\\
\hline
Agenthospital \cite{li2024agent} & \cmark & \xmark & \cmark & \cmark\\
\hline
Pharmagent \cite{gao2025pharmagents} & \xmark & \cmark & \xmark & \cmark\\
\hline
Txagent \cite{gao2025txagent} & \cmark & \dmark & \xmark & \cmark\\
\hline
Autoct \cite{liu2025autoct} & \xmark & \cmark & \xmark & \cmark\\
\hline
Kg4diagnosis \cite{zuo2025kg4diagnosis} & \dmark & \xmark & \xmark & \cmark\\
\hline
Mdteamgpt \cite{chen2025mdteamgpt} & \dmark & \xmark & \xmark & \cmark\\
\hline
Mitigating \cite{ke2024mitigating} & \xmark & \xmark & \xmark & \cmark\\
\hline
Polaris \cite{mukherjee2024polaris} & \dmark & \xmark & \cmark & \cmark\\
\hline
Rxstrategist \cite{van2024rx} & \cmark & \xmark & \xmark & \dmark\\
\hline
Clinicallab \cite{yan2024clinicallab} & \cmark & \xmark & \dmark & \cmark\\
\hline
Agenticllm \cite{sudarshan2024agentic} & \xmark & \xmark & \xmark & \dmark\\
\hline
Surgbox \cite{wu2024surgbox} & \dmark & \xmark & \dmark & \cmark\\
\hline
Aipatient \cite{yu2024simulated} & \xmark & \xmark & \cmark & \cmark\\
\hline
Healthagent \cite{abbasian2023conversational} & \xmark & \xmark & \cmark & \dmark\\
\hline
Icdcoding \cite{li2024exploring} & \dmark & \xmark & \xmark & \cmark\\
\hline
Aha \cite{pandey2024advancing} & \xmark & \xmark & \xmark & \dmark\\
\hline
Ddo \cite{jia2025ddo} & \xmark & \xmark & \cmark & \cmark\\
\hline
Beyond \cite{wang2024beyond} & \xmark & \xmark & \xmark & \cmark\\
\hline
Zodiac \cite{zhou2024zodiac} & \xmark & \xmark & \xmark & \cmark\\
\hline
Fhirviz \cite{almutairi2024fhirviz} & \xmark & \xmark & \xmark & \cmark \\
\hline
Fhir \cite{de2024multi} & \xmark & \xmark & \xmark & \cmark \\
\hline
Fuasagent \cite{zhao2025autonomous} & \cmark & \xmark & \xmark & \dmark \\
\hline
Aitherapist \cite{wasenmuller2024script} & \dmark & \xmark & \cmark & \dmark \\
\hline
Radiology \cite{zeng2024enhancing} & \xmark & \xmark & \xmark & \cmark \\
\hline
Triageagent \cite{lu2024triageagent} & \cmark & \xmark & \xmark & \cmark \\
\hline
Piors \cite{bao2024piors} & \xmark & \xmark & \cmark & \cmark \\
\hline
Selfevolve \cite{almansoori2025self} & \xmark & \xmark & \cmark & \cmark \\
\hline
Drugagent \cite{liu2024drugagent} & \xmark & \cmark & \xmark & \cmark\\
\hline
Ctagent \cite{mao2025ct} & \xmark & \xmark & \xmark & \cmark\\
\hline
Adagent \cite{hou2025adagent} & \xmark & \xmark & \xmark & \cmark\\
\hline
Drugpilot \cite{li2025drugpilot} & \dmark & \cmark & \xmark & \cmark\\
\hline
Llmadmin \cite{gebreab2024llm} & \xmark & \xmark & \dmark & \dmark\\
\hline
Swarm \cite{song2025llm} & \xmark & \cmark & \xmark & \cmark\\
\hline
Drugdiscover \cite{ock2025large} & \xmark & \cmark & \xmark & \dmark\\
\hline
Ragdrug \cite{lee2025rag} & \dmark & \cmark & \xmark & \cmark\\
\hline
Drugreason \cite{inoue2025drugagent} & \xmark & \cmark & \xmark & \cmark\\
\hline

\end{tabular}
\label{tab:task2}
\end{table}

\normalsize
\subsubsection{Drug Discovery \& Clinical Trial Design}

Drug Discovery \& Clinical Trial Design gauges an agent’s capacity to accelerate translational pipelines from target ideation to study execution. Abilities include drug-target interaction prediction, toxicity screening, hypothesis generation and lead optimization. Trial functions include protocol authoring, eligibility criteria extraction, patient-trial matching, automated feature engineering and outcome prediction. \cmark\ is warranted when either drug discovery or clinical trial design is realized. The system should experiment around task specific benchmarks, tool integrated workflows and multi-tier validation. \dmark\ is used when functionality is narrow or upstream only such as database lookups, molecular captioning or basic analytics. It does not include any form of end-to-end optimization, interaction with experimental constraints or trial support beyond templated criteria. \xmark\ is reserved for systems with no discovery or trial capabilities.

\normalsize
\subsubsection{Patient Interaction \& Monitoring}

Patient Interaction \& Monitoring characterizes an agent’s patient facing functionality across communication, monitoring and logistics. Capabilities include two-way messaging and education. It also covers symptom checkers with targeted follow-ups, adherence check-ins along with reminders, appointment scheduling, device integration for physiologic signals and personalized conversation management. \cmark\ is earned when at least one of the following is implemented: interaction, engagement, appointment management, reminders or monitoring. Evidence may include realistic dialogue behavior, symptom tracking with feedback collection, department support, real-time monitoring or stress estimation. Dynamic symptom simulation and safe handoffs also meet this bar. \dmark\ applies to limited functionality such as basic history taking, scripted exchanges or intake only prompts without personalized monitoring. \xmark\ signals no patient facing interactions, reminders, scheduling or monitoring.

\normalsize
\subsubsection{Benchmarking \& Simulation Environment}

Benchmarking \& Simulation Environment validates whether an agent is tested under rigorous and reproducible conditions that reflect clinical work. Scope includes standardized benchmarks with clear protocols and metrics. It also includes simulations that model roles, artifacts (notes, images, vitals) and multi-turn workflows. Robust studies report comparisons to strong baselines, ablations, error analyses and when appropriate, human expert review. Accordingly, \cmark\ is used for evaluation on a substantial benchmark or a clinically realistic simulation with transparent setup, baseline comparisons and ablation studies. \dmark\ is used for narrow demos or synthetic microstudies with few samples and limited metrics. \xmark\ is assigned when no meaningful benchmarking or simulation is presented.

\normalsize
\section{Empirical Findings and Discussion}

In this section, we discuss cross-cutting trends across the 7 evaluation dimensions introduced earlier. The following subsections briefly surface where healthcare LLM agents cluster. For example, the analysis reveals strong External Knowledge Integration but weak Dynamic Updates \& Forgetting within Knowledge Management, relatively prevalent Conversational Mode but sparse Error Recovery within Interaction Patterns and overall low adoption of Adaptation \& Learning. We further explain why these patterns matter for clinical reliability and operational readiness. This section then converges on critical weak points and possible actionable directions highlighting where methodological investment is most urgent.

\begin{figure*}
    \centering
    \includegraphics[width=0.5\linewidth]{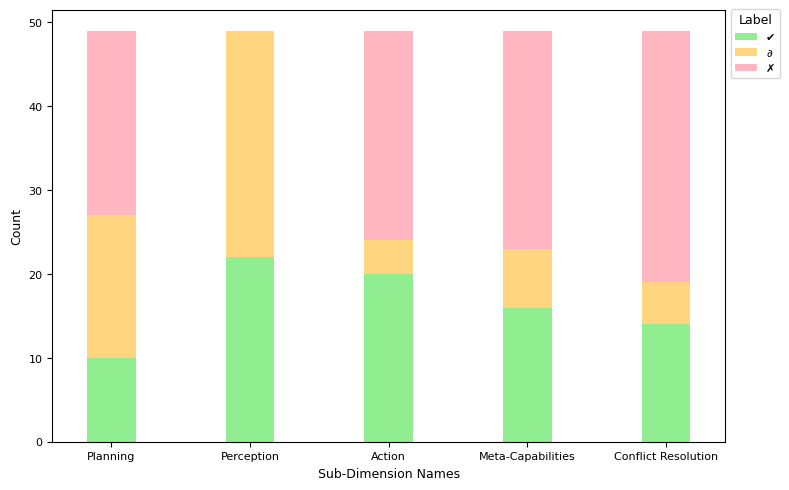}
    \caption{Distribution of Labels Across Sub-Dimensions of Cognitive Capabilities}
    \label{fig:cognitive}
\end{figure*}

\subsection{Which Cognitive Sub-Dimensions Remain Underdeveloped in Healthcare Agents?}

Figure \ref{fig:cognitive} underscores marked asymmetries in implementation prevalence across Cognitive Capabilities among surveyed LLM-based agents. \xmark\ dominates three sub-dimensions with Planning absent in $\sim$45\% of the surveyed papers, Meta-Capabilities in $\sim$53\% and Consistency \& Conflict Resolution in $\sim$61\%. This conveys that many systems still lack robust mechanisms for long horizon task decomposition, self-monitoring and systematic reconciliation of conflicting evidence. Contrarily, \cmark\ appears more frequently in Perception ($\sim$46\%) and Action ($\sim$41\%) where progress has been driven by multimodal representation learning and API or tool execution pipelines. \dmark\ occupies the middle in ways that signal partial maturity rather than stability. For instance, Perception shows $\sim$55\% \dmark\ whereas Action has only $\sim$8\% stressing that prototypes do demonstrate potential but fall short of reliable end-to-end integration. This pattern may be consequential because cognitive competencies form the foundation of trustworthy clinical agents. Without planning competence, agents risk fragmented care pathways and missed contingencies in triage or prior authorization. Weak meta-control translates into poor uncertainty calibration and absent escalation triggers. Scarce conflict resolution pipelines invite contradictory chart updates or unsafe recommendations. Advances in perception and execution while necessary, address only surface layers. Agents that faithfully ingest EHR data and generate structured outputs still fail if they cannot resolve contradictions or adapt plans to evolving evidence. These gaps suggest candidate research directions starting with formalizing planning protocols aligned to clinical workflows, continuing with embedding uncertainty aware meta-control loops and finally completing with developing contradiction detection pipelines.


\subsection{Are Knowledge Management capabilities evenly implemented across healthcare LLM agents?}

\begin{figure}
    \centering
    \includegraphics[width=0.68\linewidth]{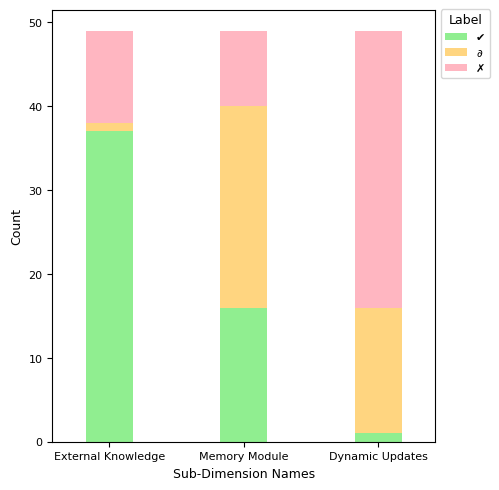}
    \caption{Distribution of Labels Across Sub-Dimensions of Knowledge Management}
    \label{fig:knowledge}
\end{figure}

Figure \ref{fig:knowledge} shows a stratified implementation pattern across three sub-dimensions including External Knowledge Integration, Memory Module and Dynamic Updates \& Forgetting. External Knowledge Integration is the most frequent with \cmark\ on 37/49 ($\sim$76\%), \dmark\ on 1/49 ($\sim$2\%) and \xmark\ on 11/49 ($\sim$22\%). These numbers emphasize widespread pairing of parametric recall with authoritative external sources at inference time. Memory Module skews \dmark\ ($\sim$49\%) with fewer \cmark\ systems ($\sim$33\%) and the remainder \xmark\ ($\sim$18\%). Many agents keep session history or simple buffers. Yet far fewer implement controller-governed, persistent, multi-tier memory with principled read/write. The most conspicuous weakness is Dynamic Updates \& Forgetting, which records just 1/49 ($\sim$2\%) \cmark\ but 33/49 ($\sim$67\%) \xmark. Very few systems focus on both sides of knowledge maintenance allowing continuous additions and explicit forgetting via decay and deprecation rules. This distribution has a high significance because knowledge currency, coherence and hygiene underpin safe clinical use. Agents must align outputs with evolving guidelines, drug recalls and formulary changes while suppressing stale or contradictory facts to prevent drift. This analysis further sheds light on the need of elevating memory from buffers to persistent substrates, coupling retrieval with multi-tier memory and enforcing deprecation detection gates.

\normalsize

\subsection{Do Current Agents Balance Conversation, Event Triggers, Human-in-the-Loop and Recovery Adequately?}

\begin{figure}
    \centering
    \includegraphics[width=0.85\linewidth]{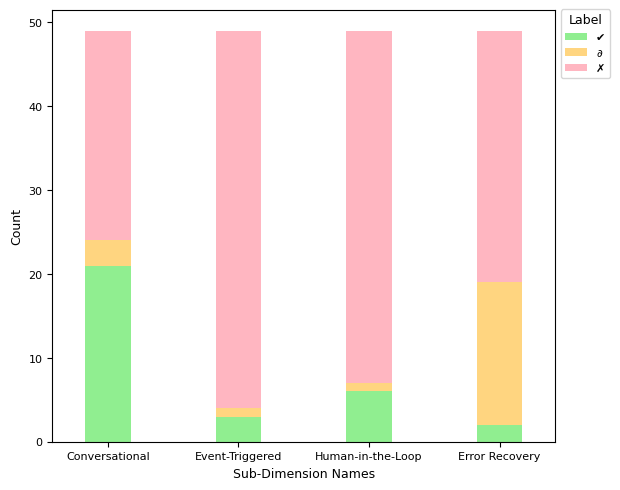}
    \caption{Distribution of Labels Across Sub-Dimensions of Interaction Patterns}
    \label{fig:interaction}
\end{figure}

Figure \ref{fig:interaction} maps implementation distribution across four Interaction Patterns sub-dimensions which are Conversational Mode, Event-Triggered Activation, Human-in-the-Loop and Error Recovery. The distribution is heterogeneous but lopsided in key areas. Conversational Mode is relatively prevalent with \cmark\ on $\sim$43\% of research works and \dmark\ on $\sim$6\% of them. This conveys that many systems support multi-turn exchange and state carryover though nearly half still fall short. Event-Triggered Activation is starkly underrepresented. It has only 3/49 ($\sim$6\%) \cmark\ and 1/49 ($\sim$2\%) \dmark\ versus 45/49 ($\sim$92\%) \xmark\ suggesting heavy reliance on manual prompts rather than external signals such as device telemetry or EHR events. Human-in-the-Loop records $\sim$12\% \cmark, $\sim$2\% \dmark\ but $\sim$86\% \xmark\ reflecting pipelines that lack mid-execution checkpoints and documented escalation paths. Error Recovery is similarly thin, with just $\sim$4\% \cmark. Only few works demonstrate idempotent tool use, rollbacks or safe mode degradation. For clinical operations, we outline potential design remedies. First, tie activation to governed event streams with provenance and debouncing. Next, define Human-in-the-Loop checkpoints as explicit gates with approval queues and uncertainty thresholds. Finally, treat recovery as a first class subsystem with bounded retries and circuit breakers. These controls can turn interactions into reviewable workflows that can in turn work toward satisfying clinical governance and reducing operational risk.

\normalsize

\subsection{How Well Do Healthcare LLM Agents Adapt?}

\begin{figure}
    \centering
    \includegraphics[width=0.68\linewidth]{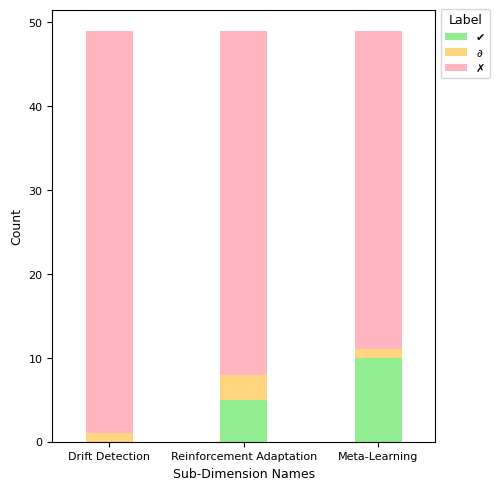}
    \caption{Distribution of Labels Across Sub-Dimensions of Adaptation \& Learning}
    \label{fig:adaptation}
\end{figure}

Figure \ref{fig:adaptation} shows uniformly low adoption across Adaptation \& Learning. \xmark\ or \textit{Not Implemented} is the prevailing label across all three sub-dimensions with only thin tails of \dmark\ and rare \cmark\ instances. Drift Detection \& Mitigation is essentially absent 0\% \cmark, $\sim$2\% (1/49) \dmark\ and $\sim$98\% (48/49) \xmark. This essentially means most agents would not notice, let alone correct shifts in codes, templates or patient mix. Reinforcement-Based Adaptation appears only sporadically with only 5 ($\sim$10\%) studies having \cmark\ and 3 ($\sim$6\%) having \dmark\ against 41 ($\sim$84\%) having \xmark\ suggesting that outcome-linked updates (rewards, constraints, clinician preferences) are rarely operationalized. Meta-Learning \& Few-Shot shows the best footing but remains far from routine with ~20\% \cmark\ and ~2\% \dmark\ but ~78\% \xmark. These numbers reveal that most systems still rely on static prompts rather than lightweight on the fly adaptation. Looking across these studies, three missing pieces recur: (i) limited lifecycle instrumentation (few pipelines report drift alarm rates or rollback metrics) (ii) underuse of preference learning and outcome-linked rewards beyond ad hoc feedback (iii) scarce end-to-end designs that fuse few-shot competence with continuous monitoring. Healthcare is non-stationary by default and without detection and controlled response, performance quietly erodes where safety and auditability matter most. In summary, today’s figures with 0-20\% \cmark\ ranges across sub-dimensions indicate islands of adaptation capability. A possible path forward is to stitch these islands into a monitored, reversible adaptation loop that can keep pace with evolving practice.

\normalsize

\subsection{Safety \& Ethics Implementation Maturity Remains Limited Among Surveyed Agents}

\begin{figure}
    \centering
    \includegraphics[width=0.85\linewidth]{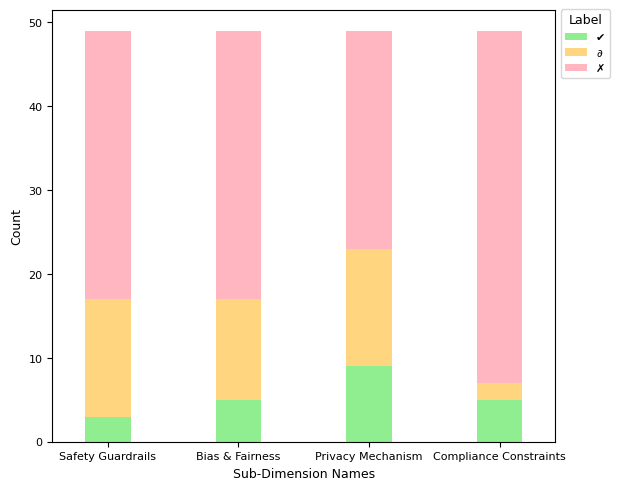}
    \caption{Distribution of Labels Across Sub-Dimensions of Safety \& Ethics}
    \label{fig:safety}
\end{figure}

Figure \ref{fig:safety} profiles four Safety \& Ethics sub-dimensions Safety Guardrails \& Adversarial Robustness, Bias \& Fairness, Privacy-Preserving Mechanisms and Regulatory \& Compliance Constraints. For Safety Guardrails \& Adversarial Robustness, only 3 out of 49 papers ($\sim$6\%) earn \cmark, with 14 ($\sim$29\%) \dmark\ and 32 ($\sim$65\%) \xmark. This tells that most agents still lack layered, testable defenses against prompt attacks or unsafe tool use. Bias \& Fairness looks similar with 5 studies ($\sim$10\%) marked \cmark, 12 with ($\sim$25\%) \dmark\ and the rest 32 ($\sim$65\%) having \xmark\ signaling that cohort-stratified evaluation and mitigation are exception rather than the norm. Privacy-Preserving Mechanisms shows the strongest footing but remains far from universal. Their numbers include $\sim$18\% \cmark, $\sim$28\% \dmark\ and $\sim$53\% \xmark. Many papers reference hosting controls, yet fewer demonstrate agent side protections with measurable leakage tests. Regulatory \& Compliance Constraints is the most underbuilt with $\sim$10\% \cmark, only $\sim$4\% \dmark\ and $\sim$86\% \xmark\ reflecting procedural assurances rarely appear as verifiable runtime gates. The results in Figure \ref{fig:safety} therefore do more than describe who implemented what. They reveal the gap between proof of concept engineering and the end-to-end assurance required for clinical adoption with 53-86\% \xmark\ across sub-dimensions. A possible remedial direction could be: (i) convert filters into layered guardrails with adversarial suites and hold-out hazard tests (ii) replace macro averages with cohort-stratified metrics (iii) elevate privacy to de-identification plus traceable access (iv) turn compliance from policy text into executable controls wired into the agent’s workflow. Closing these gaps can move the field from intent to assurance and makes safety, equity and auditability measurable. In short, the prevalence statistics in Figure \ref{fig:safety} highlight a gradient: strong conceptual intent, partial technical realization and limited end-to-end assurance. Bridging that gradient can help deliver safe, equitable and auditable LLM agents operating in real clinical workflows.

\normalsize

\subsection{Framework Typology Trends Across LLM-Based Healthcare Agents}

\normalsize
\begin{figure}
    \centering
    \includegraphics[width=0.5\linewidth]{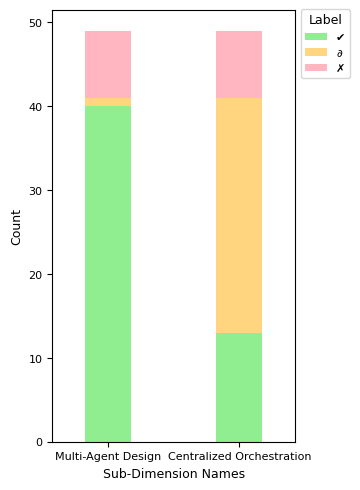}
    \caption{Distribution of Labels Across Sub-Dimensions of Framework Typology}
    \label{fig:framework}
\end{figure}

\normalsize
\begin{figure*}
    \centering
    \includegraphics[width=0.8\linewidth]{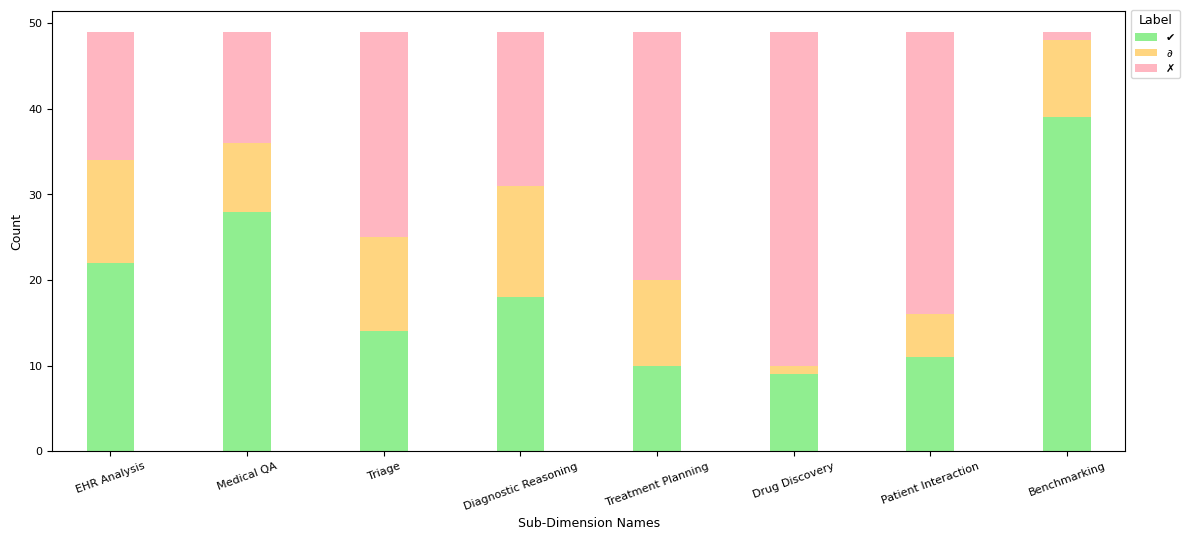}
    \caption{Distribution of Labels Across Sub-Dimensions of Core Tasks \& Subtasks}
    \label{fig:task}
\end{figure*}

Figure \ref{fig:framework} charts architectural choices within Framework Typology by tallying \cmark, \dmark\ and \xmark\ across two sub-dimensions: Multi-Agent Design and Centralized Orchestration. The data appears to have a clear asymmetry. Multi-Agent Design is broadly mature: $\sim$82\% \cmark\ with only $\sim$16\% \xmark\ indicating that role-based compositions such as planner, retriever, verifier, explainer are now the default pattern. Centralized Orchestration, meanwhile, clusters in partial territory with $\sim$57\% \dmark. This suggests that while coordination layers exist, many deployments stop short of fully realized controllers with global state, policy enforcement and auditable sequencing. This is not a verdict on what architectures ought to be but a snapshot of where engineering effort has concentrated. Teams have prioritized modular specialization allowing easier parallelization and component swaps before hardening orchestration for end-to-end assurance. In practice, typology choices shape behavior under load, evolution over time and the visibility of control. Multi-agent compositions can mirror multidisciplinary care by assigning distinct agents to guideline conformance, evidence synthesis and patient facing communication. A central orchestrator can then align these steps with institutional policy, surface uncertainty, route edge cases to clinicians and reconcile outputs prior to order. Neither pattern is universally superior. Effective systems often combine them or opt for lighter choreography when latency, cost or integration constraints make a heavyweight controller impractical. Read in this light, Figure \ref{fig:framework} indicates a field that has rapidly embraced modular, role-based design (high \cmark\ in Multi-Agent Design) while progressively layering coordination mechanisms (dominant \dmark\ in Centralized Orchestration) as deployments mature.

\subsection{Which Core Tasks Can Healthcare Agents Perform?}

Figure \ref{fig:task} depicts how often each Core Tasks \& Subtasks sub-dimension is rated \textit{Fully Implemented} \cmark, \textit{Partially Implemented} \dmark\ or \textit{Not Implemented} \xmark. The spread highlights where healthcare LLM agents already provide dependable value versus where they remain immature. Clinical Documentation \& EHR Analysis is a solid performer with 22/49 ($\sim$45\%) \cmark, 12/49 ($\sim$24\%) \dmark\ and 15/49 ($\sim$31\%) \xmark\ reflecting tractable gains in summarization and extraction. Medical Question Answering \& Decision Support is stronger at 28/49 ($\sim$57\%) \cmark, showing that retrieval-grounded recommendations are a leading use case. Conversely, Triage \& Differential Diagnosis and Diagnostic Reasoning cluster around partial adoption. Their statistics contain 14/49 ($\sim$29\%) and 18/49 ($\sim$37\%) \cmark\ alongside 11/49 ($\sim$22\%) and 13/49 ($\sim$27\%) \dmark\ respectively. This confirms their consistency with higher demands of structured hypothesis management and calibrated uncertainty. Action oriented areas remain underdeveloped. For example, Treatment Planning \& Prescription is only 10/49 ($\sim$20\%) \cmark\ and Patient Interaction \& Monitoring is 11/49 ($\sim$22\%) with both tilting heavily toward \xmark\ because of integration and safety challenges. Drug Discovery \& Clinical Trial Design records just 9/49 ($\sim$18\%) \cmark\ against 39/49 ($\sim$80\%) \xmark\ exhibiting long evaluation horizons and regulatory burden. Benchmarking \& Simulation Environment stands out with 39/49 ($\sim$80\%) \cmark\ and only 1/49 ($\sim$2\%) \xmark\ marking broad adoption of standardized datasets and increasingly realistic simulators that boost comparability and iteration speed. Overall, interpreting the numbers in Figure \ref{fig:task} shows information centric tasks like documentation, decision support and benchmarking leading in maturity. On the other hand, tasks centered around decision and action remain \textit{Partially Implemented} whereas discovery tasks lag furthest behind. This pattern maps directly to clinical workflows. Medical record summarization and question answering are ready for frontline utility but safe automation of judgment and longitudinal interaction still requires substantial governance advances.

\normalsize
\section{Conclusion}

LLM-based agents are showing early gains in automation, decision support and research, but wider clinical adoption remains limited until adaptation, safety and compliance are more robust. In this survey, we systematically review 49 research papers and propose an integrated taxonomy for empirical evaluation of these agents. Our taxonomy provides a structured lens to examine the cognitive, interactive, adaptive and ethical aspects of Agentic AI-grounded LLM-based agents along with their core tasks and framework architectures. Through this multidimensional analysis, we surface a variety of prevalent patterns showing methodological preferences, capability distributions and emerging design strategies within healthcare and medicine. While the field has made significant strides, several challenges still persist. These challenges particularly revolve around ensuring reliability, interpretability and ethical alignment in complex healthcare environments. By consolidating current methodologies and highlighting key trends, our review offers a foundational resource for researchers aiming to advance the development of robust, efficient and ethically responsible LLM-based agents. Future research can build upon this framework to explore underrepresented dimensions, refine agent capabilities and address domain specific needs. This would ultimately contribute towards more reliable, safe and effective applications of LLM-based agents in healthcare and medicine. Finally, while scaling agentic systems from research prototypes to production involves orchestration, integration, latency/cost and reliability concerns, such deployment-scale engineering lies beyond the scope of this work. We flag scalability as an open challenge and an important axis for future work but our evidence and analysis are intentionally limited to peer-reviewed/archival research papers rather than production systems.

\bibliographystyle{IEEEtran} 
\bibliography{access}    

@article{xi2025rise,
  title={The rise and potential of large language model based agents: A survey},
  author={Xi, Zhiheng and Chen, Wenxiang and Guo, Xin and He, Wei and Ding, Yiwen and Hong, Boyang and Zhang, Ming and Wang, Junzhe and Jin, Senjie and Zhou, Enyu and others},
  journal={Science China Information Sciences},
  volume={68},
  number={2},
  pages={121101},
  year={2025},
  publisher={Springer}
}

@article{schulhoff2024prompt,
  title={The prompt report: a systematic survey of prompt engineering techniques},
  author={Schulhoff, Sander and Ilie, Michael and Balepur, Nishant and Kahadze, Konstantine and Liu, Amanda and Si, Chenglei and Li, Yinheng and Gupta, Aayush and Han, HyoJung and Schulhoff, Sevien and others},
  journal={arXiv preprint arXiv:2406.06608},
  year={2024}
}

@article{gu2023systematic,
  title={A systematic survey of prompt engineering on vision-language foundation models},
  author={Gu, Jindong and Han, Zhen and Chen, Shuo and Beirami, Ahmad and He, Bailan and Zhang, Gengyuan and Liao, Ruotong and Qin, Yao and Tresp, Volker and Torr, Philip},
  journal={arXiv preprint arXiv:2307.12980},
  year={2023}
}

@inproceedings{lu2024triageagent,
  title={Triageagent: Towards better multi-agents collaborations for large language model-based clinical triage},
  author={Lu, Meng and Ho, Brandon and Ren, Dennis and Wang, Xuan},
  booktitle={Findings of the Association for Computational Linguistics: EMNLP 2024},
  pages={5747--5764},
  year={2024}
}

@article{li2024potential,
  title={Potential Multidisciplinary Use of Large Language Models for Addressing Queries in Cardio-Oncology},
  author={Li, Pengfei and Zhang, Xuejuan and Zhu, Erjia and Yu, Shijun and Sheng, Bin and Tham, Yih Chung and Wong, Tien Yin and Ji, Hongwei},
  journal={Journal of the American Heart Association},
  volume={13},
  number={6},
  pages={e033584},
  year={2024}
}

@article{zhou2024zodiac,
  title={Zodiac: A cardiologist-level llm framework for multi-agent diagnostics},
  author={Zhou, Yuan and Zhang, Peng and Song, Mengya and Zheng, Alice and Lu, Yiwen and Liu, Zhiheng and Chen, Yong and Xi, Zhaohan},
  journal={arXiv preprint arXiv:2410.02026},
  year={2024}
}

@article{shaikh2023grounding,
  title={Grounding gaps in language model generations},
  author={Shaikh, Omar and Gligori{\'c}, Kristina and Khetan, Ashna and Gerstgrasser, Matthias and Yang, Diyi and Jurafsky, Dan},
  journal={arXiv preprint arXiv:2311.09144},
  year={2023}
}

@article{mahmood2023llm,
  title={Llm-powered conversational voice assistants: Interaction patterns, opportunities, challenges, and design guidelines},
  author={Mahmood, Amama and Wang, Junxiang and Yao, Bingsheng and Wang, Dakuo and Huang, Chien-Ming},
  journal={arXiv preprint arXiv:2309.13879},
  year={2023}
}

@article{shinn2023reflexion,
  title={Reflexion: Language agents with verbal reinforcement learning},
  author={Shinn, Noah and Cassano, Federico and Gopinath, Ashwin and Narasimhan, Karthik and Yao, Shunyu},
  journal={Advances in Neural Information Processing Systems},
  volume={36},
  pages={8634--8652},
  year={2023}
}

@article{brown2020language,
  title={Language models are few-shot learners},
  author={Brown, Tom and Mann, Benjamin and Ryder, Nick and Subbiah, Melanie and Kaplan, Jared D and Dhariwal, Prafulla and Neelakantan, Arvind and Shyam, Pranav and Sastry, Girish and Askell, Amanda and others},
  journal={Advances in neural information processing systems},
  volume={33},
  pages={1877--1901},
  year={2020}
}

@inproceedings{finn2017model,
  title={Model-agnostic meta-learning for fast adaptation of deep networks},
  author={Finn, Chelsea and Abbeel, Pieter and Levine, Sergey},
  booktitle={International conference on machine learning},
  pages={1126--1135},
  year={2017},
  organization={PMLR}
}

@article{singhal2025toward,
  title={Toward expert-level medical question answering with large language models},
  author={Singhal, Karan and Tu, Tao and Gottweis, Juraj and Sayres, Rory and Wulczyn, Ellery and Amin, Mohamed and Hou, Le and Clark, Kevin and Pfohl, Stephen R and Cole-Lewis, Heather and others},
  journal={Nature Medicine},
  volume={31},
  number={3},
  pages={943--950},
  year={2025},
  publisher={Nature Publishing Group US New York}
}

@incollection{vatsal2024canprior,
  title={Can gpt improve the state of prior authorization via guideline based automated question answering?},
  author={Vatsal, Shubham and Singh, Ayush and Tafreshi, Shabnam},
  booktitle={AI for Health Equity and Fairness: Leveraging AI to Address Social Determinants of Health},
  pages={147--158},
  year={2024},
  publisher={Springer}
}

@article{vatsal2024can,
  title={Can gpt redefine medical understanding? evaluating gpt on biomedical machine reading comprehension},
  author={Vatsal, Shubham and Singh, Ayush},
  journal={arXiv preprint arXiv:2405.18682},
  year={2024}
}

@article{guo2022evaluation,
  title={Evaluation of domain generalization and adaptation on improving model robustness to temporal dataset shift in clinical medicine},
  author={Guo, Lin Lawrence and Pfohl, Stephen R and Fries, Jason and Johnson, Alistair EW and Posada, Jose and Aftandilian, Catherine and Shah, Nigam and Sung, Lillian},
  journal={Scientific reports},
  volume={12},
  number={1},
  pages={2726},
  year={2022},
  publisher={Nature Publishing Group UK London}
}

@inproceedings{wu2024autogen,
  title={Autogen: Enabling next-gen LLM applications via multi-agent conversations},
  author={Wu, Qingyun and Bansal, Gagan and Zhang, Jieyu and Wu, Yiran and Li, Beibin and Zhu, Erkang and Jiang, Li and Zhang, Xiaoyun and Zhang, Shaokun and Liu, Jiale and others},
  booktitle={First Conference on Language Modeling},
  year={2024}
}

@article{wang2022self,
  title={Self-consistency improves chain of thought reasoning in language models},
  author={Wang, Xuezhi and Wei, Jason and Schuurmans, Dale and Le, Quoc and Chi, Ed and Narang, Sharan and Chowdhery, Aakanksha and Zhou, Denny},
  journal={arXiv preprint arXiv:2203.11171},
  year={2022}
}

@article{schick2023toolformer,
  title={Toolformer: Language models can teach themselves to use tools},
  author={Schick, Timo and Dwivedi-Yu, Jane and Dess{\`\i}, Roberto and Raileanu, Roberta and Lomeli, Maria and Hambro, Eric and Zettlemoyer, Luke and Cancedda, Nicola and Scialom, Thomas},
  journal={Advances in Neural Information Processing Systems},
  volume={36},
  pages={68539--68551},
  year={2023}
}

@article{bai2022constitutional,
  title={Constitutional ai: Harmlessness from ai feedback},
  author={Bai, Yuntao and Kadavath, Saurav and Kundu, Sandipan and Askell, Amanda and Kernion, Jackson and Jones, Andy and Chen, Anna and Goldie, Anna and Mirhoseini, Azalia and McKinnon, Cameron and others},
  journal={arXiv preprint arXiv:2212.08073},
  year={2022}
}

@article{gupta2023measuring,
  title={Measuring distributional shifts in text: the advantage of language model-based embeddings},
  author={Gupta, Gyandev and Rastegarpanah, Bashir and Iyer, Amalendu and Rubin, Joshua and Kenthapadi, Krishnaram},
  journal={arXiv preprint arXiv:2312.02337},
  year={2023}
}

@article{richter2024auditing,
  title={An Auditing Test to Detect Behavioral Shift in Language Models},
  author={Richter, Leo and He, Xuanli and Minervini, Pasquale and Kusner, Matt J},
  journal={arXiv preprint arXiv:2410.19406},
  year={2024}
}

@inproceedings{abdelnabi2025get,
  title={Get my drift? catching llm task drift with activation deltas},
  author={Abdelnabi, Sahar and Fay, Aideen and Cherubin, Giovanni and Salem, Ahmed and Fritz, Mario and Paverd, Andrew},
  booktitle={2025 IEEE Conference on Secure and Trustworthy Machine Learning (SaTML)},
  pages={43--67},
  year={2025},
  organization={IEEE}
}

@article{muralidharan2024scoping,
  title={A scoping review of reporting gaps in FDA-approved AI medical devices},
  author={Muralidharan, Vijaytha and Adewale, Boluwatife Adeleye and Huang, Caroline J and Nta, Mfon Thelma and Ademiju, Peter Oluwaduyilemi and Pathmarajah, Pirunthan and Hang, Man Kien and Adesanya, Oluwafolajimi and Abdullateef, Ridwanullah Olamide and Babatunde, Abdulhammed Opeyemi and others},
  journal={NPJ Digital Medicine},
  volume={7},
  number={1},
  pages={273},
  year={2024},
  publisher={Nature Publishing Group UK London}
}

@article{van2024eu,
  title={The EU artificial intelligence act (2024): implications for healthcare},
  author={Van Kolfschooten, Hannah and Van Oirschot, Janneke},
  journal={Health Policy},
  volume={149},
  pages={105152},
  year={2024},
  publisher={Elsevier}
}

@article{rieke2020future,
  title={The future of digital health with federated learning},
  author={Rieke, Nicola and Hancox, Jonny and Li, Wenqi and Milletari, Fausto and Roth, Holger R and Albarqouni, Shadi and Bakas, Spyridon and Galtier, Mathieu N and Landman, Bennett A and Maier-Hein, Klaus and others},
  journal={NPJ digital medicine},
  volume={3},
  number={1},
  pages={119},
  year={2020},
  publisher={Nature Publishing Group UK London}
}

@article{mireshghallah2022quantifying,
  title={Quantifying privacy risks of masked language models using membership inference attacks},
  author={Mireshghallah, Fatemehsadat and Goyal, Kartik and Uniyal, Archit and Berg-Kirkpatrick, Taylor and Shokri, Reza},
  journal={arXiv preprint arXiv:2203.03929},
  year={2022}
}

@article{obermeyer2019dissecting,
  title={Dissecting racial bias in an algorithm used to manage the health of populations},
  author={Obermeyer, Ziad and Powers, Brian and Vogeli, Christine and Mullainathan, Sendhil},
  journal={Science},
  volume={366},
  number={6464},
  pages={447--453},
  year={2019},
  publisher={American Association for the Advancement of Science}
}

@article{li2023survey,
  title={A survey on fairness in large language models},
  author={Li, Yingji and Du, Mengnan and Song, Rui and Wang, Xin and Wang, Ying},
  journal={arXiv preprint arXiv:2308.10149},
  year={2023}
}

@article{gallegos2024bias,
  title={Bias and fairness in large language models: A survey},
  author={Gallegos, Isabel O and Rossi, Ryan A and Barrow, Joe and Tanjim, Md Mehrab and Kim, Sungchul and Dernoncourt, Franck and Yu, Tong and Zhang, Ruiyi and Ahmed, Nesreen K},
  journal={Computational Linguistics},
  volume={50},
  number={3},
  pages={1097--1179},
  year={2024},
  publisher={MIT Press 255 Main Street, 9th Floor, Cambridge, Massachusetts 02142, USA~…}
}

@article{purpura2025building,
  title={Building Safe GenAI Applications: An End-to-End Overview of Red Teaming for Large Language Models},
  author={Purpura, Alberto and Wadhwa, Sahil and Zymet, Jesse and Gupta, Akshay and Luo, Andy and Rad, Melissa Kazemi and Shinde, Swapnil and Sorower, Mohammad Shahed},
  journal={arXiv preprint arXiv:2503.01742},
  year={2025}
}

@article{yi2024jailbreak,
  title={Jailbreak attacks and defenses against large language models: A survey},
  author={Yi, Sibo and Liu, Yule and Sun, Zhen and Cong, Tianshuo and He, Xinlei and Song, Jiaxing and Xu, Ke and Li, Qi},
  journal={arXiv preprint arXiv:2407.04295},
  year={2024}
}

@inproceedings{chao2025jailbreaking,
  title={Jailbreaking black box large language models in twenty queries},
  author={Chao, Patrick and Robey, Alexander and Dobriban, Edgar and Hassani, Hamed and Pappas, George J and Wong, Eric},
  booktitle={2025 IEEE Conference on Secure and Trustworthy Machine Learning (SaTML)},
  pages={23--42},
  year={2025},
  organization={IEEE}
}

@article{liu2024automatic,
  title={Automatic and universal prompt injection attacks against large language models},
  author={Liu, Xiaogeng and Yu, Zhiyuan and Zhang, Yizhe and Zhang, Ning and Xiao, Chaowei},
  journal={arXiv preprint arXiv:2403.04957},
  year={2024}
}

@article{zhang2025agentorchestra,
  title={Agentorchestra: A hierarchical multi-agent framework for general-purpose task solving},
  author={Zhang, Wentao and Cui, Ce and Zhao, Yilei and Hu, Rui and Liu, Yang and Zhou, Yahui and An, Bo},
  journal={arXiv preprint arXiv:2506.12508},
  year={2025}
}

@article{pandey2024advancing,
  title={Advancing Healthcare Automation: Multi-agent system for medical necessity justification},
  author={Pandey, Himanshu and Amod, Akhil and others},
  journal={arXiv preprint arXiv:2404.17977},
  year={2024}
}

@article{yang2025agentnet,
  title={Agentnet: Decentralized evolutionary coordination for llm-based multi-agent systems},
  author={Yang, Yingxuan and Chai, Huacan and Shao, Shuai and Song, Yuanyi and Qi, Siyuan and Rui, Renting and Zhang, Weinan},
  journal={arXiv preprint arXiv:2504.00587},
  year={2025}
}

@inproceedings{wang2025colacare,
  title={Colacare: Enhancing electronic health record modeling through large language model-driven multi-agent collaboration},
  author={Wang, Zixiang and Zhu, Yinghao and Zhao, Huiya and Zheng, Xiaochen and Sui, Dehao and Wang, Tianlong and Tang, Wen and Wang, Yasha and Harrison, Ewen and Pan, Chengwei and others},
  booktitle={Proceedings of the ACM on Web Conference 2025},
  pages={2250--2261},
  year={2025}
}

@article{chen2025enhancing,
  title={Enhancing diagnostic capability with multi-agents conversational large language models},
  author={Chen, Xi and Yi, Huahui and You, Mingke and Liu, WeiZhi and Wang, Li and Li, Hairui and Zhang, Xue and Guo, Yingman and Fan, Lei and Chen, Gang and others},
  journal={NPJ digital medicine},
  volume={8},
  number={1},
  pages={159},
  year={2025},
  publisher={Nature Publishing Group UK London}
}

@article{tang2023medagents,
  title={Medagents: Large language models as collaborators for zero-shot medical reasoning},
  author={Tang, Xiangru and Zou, Anni and Zhang, Zhuosheng and Li, Ziming and Zhao, Yilun and Zhang, Xingyao and Cohan, Arman and Gerstein, Mark},
  journal={arXiv preprint arXiv:2311.10537},
  year={2023}
}

@article{han2024development,
  title={Development of a Large Language Model-based Multi-Agent Clinical Decision Support System for Korean Triage and Acuity Scale (KTAS)-Based Triage and Treatment Planning in Emergency Departments},
  author={Han, Seungjun and Choi, Wongyung},
  journal={arXiv preprint arXiv:2408.07531},
  year={2024}
}

@article{inoue2025drugagent,
  title={Drugagent: Multi-agent large language model-based reasoning for drug-target interaction prediction},
  author={Inoue, Yoshitaka and Song, Tianci and Wang, Xinling and Luna, Augustin and Fu, Tianfan},
  journal={ArXiv},
  pages={arXiv--2408},
  year={2025}
}

@article{lee2025rag,
  title={RAG-Enhanced Collaborative LLM Agents for Drug Discovery},
  author={Lee, Namkyeong and De Brouwer, Edward and Hajiramezanali, Ehsan and Biancalani, Tommaso and Park, Chanyoung and Scalia, Gabriele},
  journal={arXiv preprint arXiv:2502.17506},
  year={2025}
}

@article{ock2025large,
  title={Large Language Model Agent for Modular Task Execution in Drug Discovery},
  author={Ock, Janghoon and Meda, Radheesh Sharma and Badrinarayanan, Srivathsan and Aluru, Neha S and Chandrasekhar, Achuth and Farimani, Amir Barati},
  journal={arXiv preprint arXiv:2507.02925},
  year={2025}
}

@article{song2025llm,
  title={Llm agent swarm for hypothesis-driven drug discovery},
  author={Song, Kevin and Trotter, Andrew and Chen, Jake Y},
  journal={arXiv preprint arXiv:2504.17967},
  year={2025}
}

@inproceedings{gebreab2024llm,
  title={Llm-based framework for administrative task automation in healthcare},
  author={Gebreab, Senay A and Salah, Khaled and Jayaraman, Raja and ur Rehman, Muhammad Habib and Ellaham, Samer},
  booktitle={2024 12th International Symposium on Digital Forensics and Security (ISDFS)},
  pages={1--7},
  year={2024},
  organization={IEEE}
}

@article{li2025drugpilot,
  title={DrugPilot: LLM-based Parameterized Reasoning Agent for Drug Discovery},
  author={Li, Kun and Wu, Zhennan and Wang, Shoupeng and Wu, Jia and Pan, Shirui and Hu, Wenbin},
  journal={arXiv preprint arXiv:2505.13940},
  year={2025}
}

@article{hou2025adagent,
  title={ADAgent: LLM Agent for Alzheimer's Disease Analysis with Collaborative Coordinator},
  author={Hou, Wenlong and Yang, Guangqian and Du, Ye and Lau, Yeung and Liu, Lihao and He, Junjun and Long, Ling and Wang, Shujun},
  journal={arXiv preprint arXiv:2506.11150},
  year={2025}
}

@article{mao2025ct,
  title={CT-Agent: A Multimodal-LLM Agent for 3D CT Radiology Question Answering},
  author={Mao, Yuren and Xu, Wenyi and Qin, Yuyang and Gao, Yunjun},
  journal={arXiv preprint arXiv:2505.16229},
  year={2025}
}

@article{liu2024drugagent,
  title={Drugagent: Automating ai-aided drug discovery programming through llm multi-agent collaboration},
  author={Liu, Sizhe and Lu, Yizhou and Chen, Siyu and Hu, Xiyang and Zhao, Jieyu and Lu, Yingzhou and Zhao, Yue},
  journal={arXiv preprint arXiv:2411.15692},
  year={2024}
}

@article{almansoori2025self,
  title={Self-Evolving Multi-Agent Simulations for Realistic Clinical Interactions},
  author={Almansoori, Mohammad and Kumar, Komal and Cholakkal, Hisham},
  journal={arXiv preprint arXiv:2503.22678},
  year={2025}
}

@article{bao2024piors,
  title={Piors: Personalized intelligent outpatient reception based on large language model with multi-agents medical scenario simulation},
  author={Bao, Zhijie and Liu, Qingyun and Guo, Ying and Ye, Zhengqiang and Shen, Jun and Xie, Shirong and Peng, Jiajie and Huang, Xuanjing and Wei, Zhongyu},
  journal={arXiv preprint arXiv:2411.13902},
  year={2024}
}

@article{zeng2024enhancing,
  title={Enhancing llms for impression generation in radiology reports through a multi-agent system},
  author={Zeng, Fang and Lyu, Zhiliang and Li, Quanzheng and Li, Xiang},
  journal={arXiv preprint arXiv:2412.06828},
  year={2024}
}

@article{wasenmuller2024script,
  title={Script-Based Dialog Policy Planning for LLM-Powered Conversational Agents: A Basic Architecture for an" AI Therapist"},
  author={Wasenm{\"u}ller, Robert and Hilbert, Kevin and Benzm{\"u}ller, Christoph},
  journal={arXiv preprint arXiv:2412.15242},
  year={2024}
}

@article{zhao2025autonomous,
  title={Autonomous Multi-Modal LLM Agents for Treatment Planning in Focused Ultrasound Ablation Surgery},
  author={Zhao, Lina and Bai, Jiaxing and Bian, Zihao and Chen, Qingyue and Li, Yafang and Li, Guangbo and He, Min and Yao, Huaiyuan and Zhang, Zongjiu},
  journal={arXiv preprint arXiv:2505.21418},
  year={2025}
}

@inproceedings{de2024multi,
  title={A multi-agent architecture for privacy-preserving natural language interaction with fhir-based electronic health records},
  author={De Maio, Carmen and Fenza, Giuseppe and Furno, Domenico and Grauso, Teodoro and Loia, Vincenzo},
  booktitle={2024 International Conference on Software, Telecommunications and Computer Networks (SoftCOM)},
  pages={1--6},
  year={2024},
  organization={IEEE}
}

@inproceedings{almutairi2024fhirviz,
  title={Fhirviz: Multi-agent platform for fhir visualization to advance healthcare analytics},
  author={ALMutairi, Mariam and AlKulaib, Lulwah and Wang, Shengkun and Chen, Zhiqian and ALMutairi, Youssif and Alenazi, Thamer M and Luther, Kurt and Lu, Chang-Tien},
  booktitle={Proceedings of the 15th ACM International Conference on Bioinformatics, Computational Biology and Health Informatics},
  pages={1--7},
  year={2024}
}

@article{wang2024beyond,
  title={Beyond direct diagnosis: LLM-based multi-specialist agent consultation for automatic diagnosis},
  author={Wang, Haochun and Zhao, Sendong and Qiang, Zewen and Xi, Nuwa and Qin, Bing and Liu, Ting},
  journal={arXiv preprint arXiv:2401.16107},
  year={2024}
}

@article{jia2025ddo,
  title={DDO: Dual-Decision Optimization via Multi-Agent Collaboration for LLM-Based Medical Consultation},
  author={Jia, Zhihao and Jia, Mingyi and Duan, Junwen and Wang, Jianxin},
  journal={arXiv preprint arXiv:2505.18630},
  year={2025}
}

@article{li2024exploring,
  title={Exploring llm multi-agents for icd coding},
  author={Li, Rumeng and Wang, Xun and Yu, Hong},
  journal={arXiv preprint arXiv:2406.15363},
  year={2024}
}

@article{abbasian2023conversational,
  title={Conversational health agents: A personalized llm-powered agent framework},
  author={Abbasian, Mahyar and Azimi, Iman and Rahmani, Amir M and Jain, Ramesh},
  journal={arXiv preprint arXiv:2310.02374},
  year={2023}
}

@article{yu2024simulated,
  title={Simulated patient systems are intelligent when powered by large language model-based AI agents},
  author={Yu, Huizi and Zhou, Jiayan and Li, Lingyao and Chen, Shan and Gallifant, Jack and Shi, Anye and Li, Xiang and He, Jingxian and Hua, Wenyue and Jin, Mingyu and others},
  journal={arXiv preprint arXiv:2409.18924},
  year={2024}
}

@inproceedings{wu2024surgbox,
  title={Surgbox: Agent-driven operating room sandbox with surgery copilot},
  author={Wu, Jinlin and Liang, Xusheng and Bai, Xuexue and Chen, Zhen},
  booktitle={2024 IEEE International Conference on Big Data (BigData)},
  pages={2041--2048},
  year={2024},
  organization={IEEE}
}

@article{sudarshan2024agentic,
  title={Agentic llm workflows for generating patient-friendly medical reports},
  author={Sudarshan, Malavikha and Shih, Sophie and Yee, Estella and Yang, Alina and Zou, John and Chen, Cathy and Zhou, Quan and Chen, Leon and Singhal, Chinmay and Shih, George},
  journal={arXiv preprint arXiv:2408.01112},
  year={2024}
}

@article{yan2024clinicallab,
  title={Clinicallab: Aligning agents for multi-departmental clinical diagnostics in the real world},
  author={Yan, Weixiang and Liu, Haitian and Wu, Tengxiao and Chen, Qian and Wang, Wen and Chai, Haoyuan and Wang, Jiayi and Zhao, Weishan and Zhang, Yixin and Zhang, Renjun and others},
  journal={arXiv preprint arXiv:2406.13890},
  year={2024}
}

@article{van2024rx,
  title={Rx strategist: Prescription verification using llm agents system},
  author={Van, Phuc Phan and Minh, Dat Nguyen and Ngoc, An Dinh and Thanh, Huy Phan},
  journal={arXiv preprint arXiv:2409.03440},
  year={2024}
}

@article{asai2024self,
  title={Self-rag: Learning to retrieve, generate, and critique through self-reflection},
  author={Asai, Akari and Wu, Zeqiu and Wang, Yizhong and Sil, Avirup and Hajishirzi, Hannaneh},
  year={2024},
  publisher={ICLR}
}

@article{alayrac2022flamingo,
  title={Flamingo: a visual language model for few-shot learning},
  author={Alayrac, Jean-Baptiste and Donahue, Jeff and Luc, Pauline and Miech, Antoine and Barr, Iain and Hasson, Yana and Lenc, Karel and Mensch, Arthur and Millican, Katherine and Reynolds, Malcolm and others},
  journal={Advances in neural information processing systems},
  volume={35},
  pages={23716--23736},
  year={2022}
}

@article{meng2022mass,
  title={Mass-editing memory in a transformer},
  author={Meng, Kevin and Sharma, Arnab Sen and Andonian, Alex and Belinkov, Yonatan and Bau, David},
  journal={arXiv preprint arXiv:2210.07229},
  year={2022}
}

@article{shi2024enhancing,
  title={Enhancing Clinical Trial Patient Matching through Knowledge Augmentation and Reasoning with Multi-Agents},
  author={Shi, Hanwen and Zhang, Jin and Zhang, Kunpeng},
  journal={arXiv preprint arXiv:2411.14637},
  year={2024}
}

@article{khandelwal2019generalization,
  title={Generalization through memorization: Nearest neighbor language models},
  author={Khandelwal, Urvashi and Levy, Omer and Jurafsky, Dan and Zettlemoyer, Luke and Lewis, Mike},
  journal={arXiv preprint arXiv:1911.00172},
  year={2019}
}

@article{jiang2025medagentbench,
  title={Medagentbench: A realistic virtual ehr environment to benchmark medical llm agents},
  author={Jiang, Yixing and Black, Kameron C and Geng, Gloria and Park, Danny and Zou, James and Ng, Andrew Y and Chen, Jonathan H},
  journal={arXiv preprint arXiv:2501.14654},
  year={2025}
}

@article{schmidgall2024agentclinic,
  title={AgentClinic: a multimodal agent benchmark to evaluate AI in simulated clinical environments},
  author={Schmidgall, Samuel and Ziaei, Rojin and Harris, Carl and Reis, Eduardo and Jopling, Jeffrey and Moor, Michael},
  journal={arXiv preprint arXiv:2405.07960},
  year={2024}
}

@article{merrill2024transforming,
  title={Transforming wearable data into health insights using large language model agents},
  author={Merrill, Mike A and Paruchuri, Akshay and Rezaei, Naghmeh and Kovacs, Geza and Perez, Javier and Liu, Yun and Schenck, Erik and Hammerquist, Nova and Sunshine, Jake and Tailor, Shyam and others},
  journal={arXiv preprint arXiv:2406.06464},
  year={2024}
}

@article{jin2024matching,
  title={Matching patients to clinical trials with large language models},
  author={Jin, Qiao and Wang, Zifeng and Floudas, Charalampos S and Chen, Fangyuan and Gong, Changlin and Bracken-Clarke, Dara and Xue, Elisabetta and Yang, Yifan and Sun, Jimeng and Lu, Zhiyong},
  journal={Nature communications},
  volume={15},
  number={1},
  pages={9074},
  year={2024},
  publisher={Nature Publishing Group UK London}
}

@article{m2024augmenting,
  title={Augmenting large language models with chemistry tools},
  author={M. Bran, Andres and Cox, Sam and Schilter, Oliver and Baldassari, Carlo and White, Andrew D and Schwaller, Philippe},
  journal={Nature Machine Intelligence},
  volume={6},
  number={5},
  pages={525--535},
  year={2024},
  publisher={Nature Publishing Group UK London}
}

@article{deo2025conversational,
  title={A conversational artificial intelligence agent for medication reconciliation and review},
  author={Deo, Rahul C and Goto, Shinichi and Jain, Tara and Meier, Sarah and Patel, Rahul},
  journal={medRxiv},
  pages={2025--06},
  year={2025},
  publisher={Cold Spring Harbor Laboratory Press}
}

@inproceedings{shi2024ehragent,
  title={Ehragent: Code empowers large language models for few-shot complex tabular reasoning on electronic health records},
  author={Shi, Wenqi and Xu, Ran and Zhuang, Yuchen and Yu, Yue and Zhang, Jieyu and Wu, Hang and Zhu, Yuanda and Ho, Joyce and Yang, Carl and Wang, May D},
  booktitle={Proceedings of the Conference on Empirical Methods in Natural Language Processing. Conference on Empirical Methods in Natural Language Processing},
  volume={2024},
  pages={22315},
  year={2024}
}

@article{mukherjee2024polaris,
  title={Polaris: A safety-focused llm constellation architecture for healthcare},
  author={Mukherjee, Subhabrata and Gamble, Paul and Ausin, Markel Sanz and Kant, Neel and Aggarwal, Kriti and Manjunath, Neha and Datta, Debajyoti and Liu, Zhengliang and Ding, Jiayuan and Busacca, Sophia and others},
  journal={arXiv preprint arXiv:2403.13313},
  year={2024}
}

@inproceedings{cai2019human,
  title={Human-centered tools for coping with imperfect algorithms during medical decision-making},
  author={Cai, Carrie J and Reif, Emily and Hegde, Narayan and Hipp, Jason and Kim, Been and Smilkov, Daniel and Wattenberg, Martin and Viegas, Fernanda and Corrado, Greg S and Stumpe, Martin C and others},
  booktitle={Proceedings of the 2019 chi conference on human factors in computing systems},
  pages={1--14},
  year={2019}
}

@article{ke2024mitigating,
  title={Mitigating cognitive biases in clinical decision-making through multi-agent conversations using large language models: simulation study},
  author={Ke, Yuhe and Yang, Rui and Lie, Sui An and Lim, Taylor Xin Yi and Ning, Yilin and Li, Irene and Abdullah, Hairil Rizal and Ting, Daniel Shu Wei and Liu, Nan},
  journal={Journal of Medical Internet Research},
  volume={26},
  pages={e59439},
  year={2024},
  publisher={JMIR Publications Toronto, Canada}
}

@article{rose2025meddxagent,
  title={Meddxagent: A unified modular agent framework for explainable automatic differential diagnosis},
  author={Rose, Daniel and Hung, Chia-Chien and Lepri, Marco and Alqassem, Israa and Gashteovski, Kiril and Lawrence, Carolin},
  journal={arXiv preprint arXiv:2502.19175},
  year={2025}
}

@article{chen2025mdteamgpt,
  title={Mdteamgpt: A self-evolving llm-based multi-agent framework for multi-disciplinary team medical consultation},
  author={Chen, Kai and Li, Xinfeng and Yang, Tianpei and Wang, Hewei and Dong, Wei and Gao, Yang},
  journal={arXiv preprint arXiv:2503.13856},
  year={2025}
}

@inproceedings{zuo2025kg4diagnosis,
  title={Kg4diagnosis: A hierarchical multi-agent llm framework with knowledge graph enhancement for medical diagnosis},
  author={Zuo, Kaiwen and Jiang, Yirui and Mo, Fan and Lio, Pietro},
  booktitle={AAAI Bridge Program on AI for Medicine and Healthcare},
  pages={195--204},
  year={2025},
  organization={PMLR}
}

@article{liu2025autoct,
  title={AUTOCT: Automating Interpretable Clinical Trial Prediction with LLM Agents},
  author={Liu, Fengze and Wang, Haoyu and Cho, Joonhyuk and Roth, Dan and Lo, Andrew W},
  journal={arXiv preprint arXiv:2506.04293},
  year={2025}
}

@article{gao2025txagent,
  title={TxAgent: An AI agent for therapeutic reasoning across a universe of tools},
  author={Gao, Shanghua and Zhu, Richard and Kong, Zhenglun and Noori, Ayush and Su, Xiaorui and Ginder, Curtis and Tsiligkaridis, Theodoros and Zitnik, Marinka},
  journal={arXiv preprint arXiv:2503.10970},
  year={2025}
}

@article{gao2025pharmagents,
  title={Pharmagents: Building a virtual pharma with large language model agents},
  author={Gao, Bowen and Huang, Yanwen and Liu, Yiqiao and Xie, Wenxuan and Ma, Wei-Ying and Zhang, Ya-Qin and Lan, Yanyan},
  journal={arXiv preprint arXiv:2503.22164},
  year={2025}
}

@article{li2024agent,
  title={Agent hospital: A simulacrum of hospital with evolvable medical agents},
  author={Li, Junkai and Lai, Yunghwei and Li, Weitao and Ren, Jingyi and Zhang, Meng and Kang, Xinhui and Wang, Siyu and Li, Peng and Zhang, Ya-Qin and Ma, Weizhi and others},
  journal={arXiv preprint arXiv:2405.02957},
  year={2024}
}

@article{kim2024mdagents,
  title={Mdagents: An adaptive collaboration of llms for medical decision-making},
  author={Kim, Yubin and Park, Chanwoo and Jeong, Hyewon and Chan, Yik S and Xu, Xuhai and McDuff, Daniel and Lee, Hyeonhoon and Ghassemi, Marzyeh and Breazeal, Cynthia and Park, Hae W},
  journal={Advances in Neural Information Processing Systems},
  volume={37},
  pages={79410--79452},
  year={2024}
}

@article{jin2024agentmd,
  title={Agentmd: Empowering language agents for risk prediction with large-scale clinical tool learning},
  author={Jin, Qiao and Wang, Zhizheng and Yang, Yifan and Zhu, Qingqing and Wright, Donald and Huang, Thomas and Wilbur, W John and He, Zhe and Taylor, Andrew and Chen, Qingyu and others},
  journal={arXiv preprint arXiv:2402.13225},
  year={2024}
}

@article{yang2024medaide,
  title={MedAide: Information Fusion and Anatomy of Medical Intents via LLM-based Agent Collaboration},
  author={Yang, Dingkang and Wei, Jinjie and Li, Mingcheng and Liu, Jiyao and Liu, Lihao and Hu, Ming and He, Junjun and Ju, Yakun and Zhou, Wei and Liu, Yang and others},
  journal={arXiv e-prints},
  pages={arXiv--2410},
  year={2024}
}

@article{yao2023tree,
  title={Tree of thoughts: Deliberate problem solving with large language models, 2023},
  author={Yao, Shunyu and Yu, Dian and Zhao, Jeffrey and Shafran, Izhak and Griffiths, Thomas L and Cao, Yuan and Narasimhan, Karthik},
  journal={URL https://arxiv. org/abs/2305.10601},
  volume={3},
  pages={1},
  year={2023}
}

@article{chowdhery2023palm,
  title={Palm: Scaling language modeling with pathways},
  author={Chowdhery, Aakanksha and Narang, Sharan and Devlin, Jacob and Bosma, Maarten and Mishra, Gaurav and Roberts, Adam and Barham, Paul and Chung, Hyung Won and Sutton, Charles and Gehrmann, Sebastian and others},
  journal={Journal of Machine Learning Research},
  volume={24},
  number={240},
  pages={1--113},
  year={2023}
}

@article{baker2024chatgpt,
  title={ChatGPT's ability to assist with clinical documentation: a randomized controlled trial},
  author={Baker, Hayden P and Dwyer, Emma and Kalidoss, Senthooran and Hynes, Kelly and Wolf, Jennifer and Strelzow, Jason A},
  journal={JAAOS-Journal of the American Academy of Orthopaedic Surgeons},
  volume={32},
  number={3},
  pages={123--129},
  year={2024},
  publisher={LWW}
}

@inproceedings{leong2024efficient,
  title={Efficient fine-tuning of large language models for automated medical documentation},
  author={Leong, Huiyi and Gao, Yifan and Ji, Shuai and Zhang, Yang and Pamuksuz, Uktu},
  booktitle={2024 4th International Conference on Digital Society and Intelligent Systems (DSInS)},
  pages={204--209},
  year={2024},
  organization={IEEE}
}

@article{subramanian2024enhancing,
  title={Enhancing health care communication with large language models—the role, challenges, and future directions},
  author={Subramanian, Charumathi Raghu and Yang, Daniel A and Khanna, Raman},
  journal={JAMA network open},
  volume={7},
  number={3},
  pages={e240347--e240347},
  year={2024},
  publisher={American Medical Association}
}

@article{chakraborty2023artificial,
  title={Artificial intelligence enabled ChatGPT and large language models in drug target discovery, drug discovery, and development},
  author={Chakraborty, Chiranjib and Bhattacharya, Manojit and Lee, Sang-Soo},
  journal={Molecular therapy Nucleic acids},
  volume={33},
  pages={866--868},
  year={2023},
  publisher={Elsevier}
}

@article{singhal2023large,
  title={Large language models encode clinical knowledge},
  author={Singhal, Karan and Azizi, Shekoofeh and Tu, Tao and Mahdavi, S Sara and Wei, Jason and Chung, Hyung Won and Scales, Nathan and Tanwani, Ajay and Cole-Lewis, Heather and Pfohl, Stephen and others},
  journal={Nature},
  volume={620},
  number={7972},
  pages={172--180},
  year={2023},
  publisher={Nature Publishing Group}
}

@article{achiam2023gpt,
  title={Gpt-4 technical report},
  author={Achiam, Josh and Adler, Steven and Agarwal, Sandhini and Ahmad, Lama and Akkaya, Ilge and Aleman, Florencia Leoni and Almeida, Diogo and Altenschmidt, Janko and Altman, Sam and Anadkat, Shyamal and others},
  journal={arXiv preprint arXiv:2303.08774},
  year={2023}
}

@article{wei2022chain,
  title={Chain-of-thought prompting elicits reasoning in large language models},
  author={Wei, Jason and Wang, Xuezhi and Schuurmans, Dale and Bosma, Maarten and Xia, Fei and Chi, Ed and Le, Quoc V and Zhou, Denny and others},
  journal={Advances in neural information processing systems},
  volume={35},
  pages={24824--24837},
  year={2022}
}

@article{yan2024practical,
  title={Practical and ethical challenges of large language models in education: A systematic scoping review},
  author={Yan, Lixiang and Sha, Lele and Zhao, Linxuan and Li, Yuheng and Martinez-Maldonado, Roberto and Chen, Guanliang and Li, Xinyu and Jin, Yueqiao and Ga{\v{s}}evi{\'c}, Dragan},
  journal={British Journal of Educational Technology},
  volume={55},
  number={1},
  pages={90--112},
  year={2024},
  publisher={Wiley Online Library}
}

@inproceedings{li2023large,
  title={Large language models in finance: A survey},
  author={Li, Yinheng and Wang, Shaofei and Ding, Han and Chen, Hang},
  booktitle={Proceedings of the fourth ACM international conference on AI in finance},
  pages={374--382},
  year={2023}
}

@article{dhuliawala2023chain,
  title={Chain-of-verification reduces hallucination in large language models},
  author={Dhuliawala, Shehzaad and Komeili, Mojtaba and Xu, Jing and Raileanu, Roberta and Li, Xian and Celikyilmaz, Asli and Weston, Jason},
  journal={arXiv preprint arXiv:2309.11495},
  year={2023}
}

@article{manakul2023selfcheckgpt,
  title={Selfcheckgpt: Zero-resource black-box hallucination detection for generative large language models},
  author={Manakul, Potsawee and Liusie, Adian and Gales, Mark JF},
  journal={arXiv preprint arXiv:2303.08896},
  year={2023}
}

@article{packer2023memgpt,
  title={MemGPT: Towards LLMs as Operating Systems.},
  author={Packer, Charles and Fang, Vivian and Patil, Shishir\_G and Lin, Kevin and Wooders, Sarah and Gonzalez, Joseph\_E},
  year={2023},
  publisher={ArXiv}
}

@inproceedings{guu2020retrieval,
  title={Retrieval augmented language model pre-training},
  author={Guu, Kelvin and Lee, Kenton and Tung, Zora and Pasupat, Panupong and Chang, Mingwei},
  booktitle={International conference on machine learning},
  pages={3929--3938},
  year={2020},
  organization={PMLR}
}

@article{lewis2020retrieval,
  title={Retrieval-augmented generation for knowledge-intensive nlp tasks},
  author={Lewis, Patrick and Perez, Ethan and Piktus, Aleksandra and Petroni, Fabio and Karpukhin, Vladimir and Goyal, Naman and K{\"u}ttler, Heinrich and Lewis, Mike and Yih, Wen-tau and Rockt{\"a}schel, Tim and others},
  journal={Advances in neural information processing systems},
  volume={33},
  pages={9459--9474},
  year={2020}
}

@article{wang2023plan,
  title={Plan-and-solve prompting: Improving zero-shot chain-of-thought reasoning by large language models},
  author={Wang, Lei and Xu, Wanyu and Lan, Yihuai and Hu, Zhiqiang and Lan, Yunshi and Lee, Roy Ka-Wei and Lim, Ee-Peng},
  journal={arXiv preprint arXiv:2305.04091},
  year={2023}
}

@inproceedings{yue2024clinicalagent,
  title={Clinicalagent: Clinical trial multi-agent system with large language model-based reasoning},
  author={Yue, Ling and Xing, Sixue and Chen, Jintai and Fu, Tianfan},
  booktitle={Proceedings of the 15th ACM International Conference on Bioinformatics, Computational Biology and Health Informatics},
  pages={1--10},
  year={2024}
}

@inproceedings{yao2023react,
  title={React: Synergizing reasoning and acting in language models},
  author={Yao, Shunyu and Zhao, Jeffrey and Yu, Dian and Du, Nan and Shafran, Izhak and Narasimhan, Karthik and Cao, Yuan},
  booktitle={International Conference on Learning Representations (ICLR)},
  year={2023}
}

@article{wang2025medagent,
  title={MedAgent-Pro: Towards Evidence-Based Multi-Modal Medical Diagnosis via Reasoning Agentic Workflow},
  author={Wang, Ziyue and Wu, Junde and Cai, Linghan and Low, Chang Han and Yang, Xihong and Li, Qiaxuan and Jin, Yueming},
  journal={arXiv preprint arXiv:2503.18968},
  year={2025}
}

@inproceedings{wei2024medco,
  title={Medco: Medical education copilots based on a multi-agent framework},
  author={Wei, Hao and Qiu, Jianing and Yu, Haibao and Yuan, Wu},
  booktitle={European Conference on Computer Vision},
  pages={119--135},
  year={2024},
  organization={Springer}
}

@article{liu2025medchat,
  title={MedChat: A Multi-Agent Framework for Multimodal Diagnosis with Large Language Models},
  author={Liu, Philip R and Bansal, Sparsh and Dinh, Jimmy and Pawar, Aditya and Satishkumar, Ramani and Desai, Shail and Gupta, Neeraj and Wang, Xin and Hu, Shu},
  journal={arXiv preprint arXiv:2506.07400},
  year={2025}
}

@article{li2024mmedagent,
  title={Mmedagent: Learning to use medical tools with multi-modal agent},
  author={Li, Binxu and Yan, Tiankai and Pan, Yuanting and Luo, Jie and Ji, Ruiyang and Ding, Jiayuan and Xu, Zhe and Liu, Shilong and Dong, Haoyu and Lin, Zihao and others},
  journal={arXiv preprint arXiv:2407.02483},
  year={2024}
}

@article{fan2024ai,
  title={Ai hospital: Benchmarking large language models in a multi-agent medical interaction simulator},
  author={Fan, Zhihao and Tang, Jialong and Chen, Wei and Wang, Siyuan and Wei, Zhongyu and Xi, Jun and Huang, Fei and Zhou, Jingren},
  journal={arXiv preprint arXiv:2402.09742},
  year={2024}
}

@article{feng2025doctoragent,
  title={DoctorAgent-RL: A Multi-Agent Collaborative Reinforcement Learning System for Multi-Turn Clinical Dialogue},
  author={Feng, Yichun and Wang, Jiawei and Zhou, Lu and Li, Yixue},
  journal={arXiv preprint arXiv:2505.19630},
  year={2025}
}

@article{yehudai2025survey,
  title={Survey on evaluation of llm-based agents},
  author={Yehudai, Asaf and Eden, Lilach and Li, Alan and Uziel, Guy and Zhao, Yilun and Bar-Haim, Roy and Cohan, Arman and Shmueli-Scheuer, Michal},
  journal={arXiv preprint arXiv:2503.16416},
  year={2025}
}

@article{zhang2404survey,
  title={A survey on the memory mechanism of large language model based agents, 2024},
  author={Zhang, Zeyu and Bo, Xiaohe and Ma, Chen and Li, Rui and Chen, Xu and Dai, Quanyu and Zhu, Jieming and Dong, Zhenhua and Wen, Ji-Rong},
  journal={URL https://arxiv. org/abs/2404.13501}
}

@article{huang2024understanding,
  title={Understanding the planning of LLM agents: A survey},
  author={Huang, Xu and Liu, Weiwen and Chen, Xiaolong and Wang, Xingmei and Wang, Hao and Lian, Defu and Wang, Yasheng and Tang, Ruiming and Chen, Enhong},
  journal={arXiv preprint arXiv:2402.02716},
  year={2024}
}

@inproceedings{li2025review,
  title={A review of prominent paradigms for llm-based agents: Tool use, planning (including rag), and feedback learning},
  author={Li, Xinzhe},
  booktitle={Proceedings of the 31st International Conference on Computational Linguistics},
  pages={9760--9779},
  year={2025}
}

@article{yu2025survey,
  title={A survey on trustworthy llm agents: Threats and countermeasures},
  author={Yu, Miao and Meng, Fanci and Zhou, Xinyun and Wang, Shilong and Mao, Junyuan and Pang, Linsey and Chen, Tianlong and Wang, Kun and Li, Xinfeng and Zhang, Yongfeng and others},
  journal={arXiv preprint arXiv:2503.09648},
  year={2025}
}

@article{wang2025survey,
  title={A survey of llm-based agents in medicine: How far are we from baymax?},
  author={Wang, Wenxuan and Ma, Zizhan and Wang, Zheng and Wu, Chenghan and Ji, Jiaming and Chen, Wenting and Li, Xiang and Yuan, Yixuan},
  journal={arXiv preprint arXiv:2502.11211},
  year={2025}
}

@article{zhao2023depth,
  title={An In-depth Survey of Large Language Model-based Artificial Intelligence Agents. arXiv (2023)},
  author={Zhao, Pengyu and Jin, Zijian and Cheng, Ning},
  journal={arXiv preprint arXiv:2309.14365},
  year={2023}
}

@article{li2024survey,
  title={A survey on LLM-based multi-agent systems: workflow, infrastructure, and challenges},
  author={Li, Xinyi and Wang, Sai and Zeng, Siqi and Wu, Yu and Yang, Yi},
  journal={Vicinagearth},
  volume={1},
  number={1},
  pages={9},
  year={2024},
  publisher={Springer}
}

@inproceedings{dong2024survey,
  title={A survey of llm-based agents: Theories, technologies, applications and suggestions},
  author={Dong, Xiaofei and Zhang, Xueqiang and Bu, Weixin and Zhang, Dan and Cao, Feng},
  booktitle={2024 3rd International Conference on Artificial Intelligence, Internet of Things and Cloud Computing Technology (AIoTC)},
  pages={407--413},
  year={2024},
  organization={IEEE}
}

@article{plaat2025agentic,
  title={Agentic large language models, a survey},
  author={Plaat, Aske and van Duijn, Max and van Stein, Niki and Preuss, Mike and van der Putten, Peter and Batenburg, Kees Joost},
  journal={arXiv preprint arXiv:2503.23037},
  year={2025}
}

@article{guo2024large,
  title={Large language model based multi-agents: A survey of progress and challenges},
  author={Guo, Taicheng and Chen, Xiuying and Wang, Yaqi and Chang, Ruidi and Pei, Shichao and Chawla, Nitesh V and Wiest, Olaf and Zhang, Xiangliang},
  journal={arXiv preprint arXiv:2402.01680},
  year={2024}
}

@article{ferrag2025llm,
  title={From llm reasoning to autonomous ai agents: A comprehensive review},
  author={Ferrag, Mohamed Amine and Tihanyi, Norbert and Debbah, Merouane},
  journal={arXiv preprint arXiv:2504.19678},
  year={2025}
}

@article{nori2023capabilities,
  title={Capabilities of gpt-4 on medical challenge problems},
  author={Nori, Harsha and King, Nicholas and McKinney, Scott Mayer and Carignan, Dean and Horvitz, Eric},
  journal={arXiv preprint arXiv:2303.13375},
  year={2023}
}

@article{wang2024interactive,
  title={Interactive computer-aided diagnosis on medical image using large language models},
  author={Wang, Sheng and Zhao, Zihao and Ouyang, Xi and Liu, Tianming and Wang, Qian and Shen, Dinggang},
  journal={Communications Engineering},
  volume={3},
  number={1},
  pages={133},
  year={2024},
  publisher={Nature Publishing Group UK London}
}

@article{sahoo2024systematic,
  title={A systematic survey of prompt engineering in large language models: Techniques and applications},
  author={Sahoo, Pranab and Singh, Ayush Kumar and Saha, Sriparna and Jain, Vinija and Mondal, Samrat and Chadha, Aman},
  journal={arXiv preprint arXiv:2402.07927},
  year={2024}
}

@article{vatsal2024survey,
  title={A survey of prompt engineering methods in large language models for different nlp tasks},
  author={Vatsal, Shubham and Dubey, Harsh},
  journal={arXiv preprint arXiv:2407.12994},
  year={2024}
}

@article{vatsal2025multilingual,
  title={Multilingual Prompt Engineering in Large Language Models: A Survey Across NLP Tasks},
  author={Vatsal, Shubham and Dubey, Harsh and Singh, Aditi},
  journal={arXiv preprint arXiv:2505.11665},
  year={2025}
}

@article{yang2023large,
  title={Large language models in health care: Development, applications, and challenges},
  author={Yang, Rui and Tan, Ting Fang and Lu, Wei and Thirunavukarasu, Arun James and Ting, Daniel Shu Wei and Liu, Nan},
  journal={Health Care Science},
  volume={2},
  number={4},
  pages={255--263},
  year={2023},
  publisher={Wiley Online Library}
}

@article{thirunavukarasu2023large,
  title={Large language models in medicine},
  author={Thirunavukarasu, Arun James and Ting, Darren Shu Jeng and Elangovan, Kabilan and Gutierrez, Laura and Tan, Ting Fang and Ting, Daniel Shu Wei},
  journal={Nature medicine},
  volume={29},
  number={8},
  pages={1930--1940},
  year={2023},
  publisher={Nature Publishing Group US New York}
}

\begin{IEEEbiography}[{\includegraphics[width=1in,height=1.25in,clip,keepaspectratio]{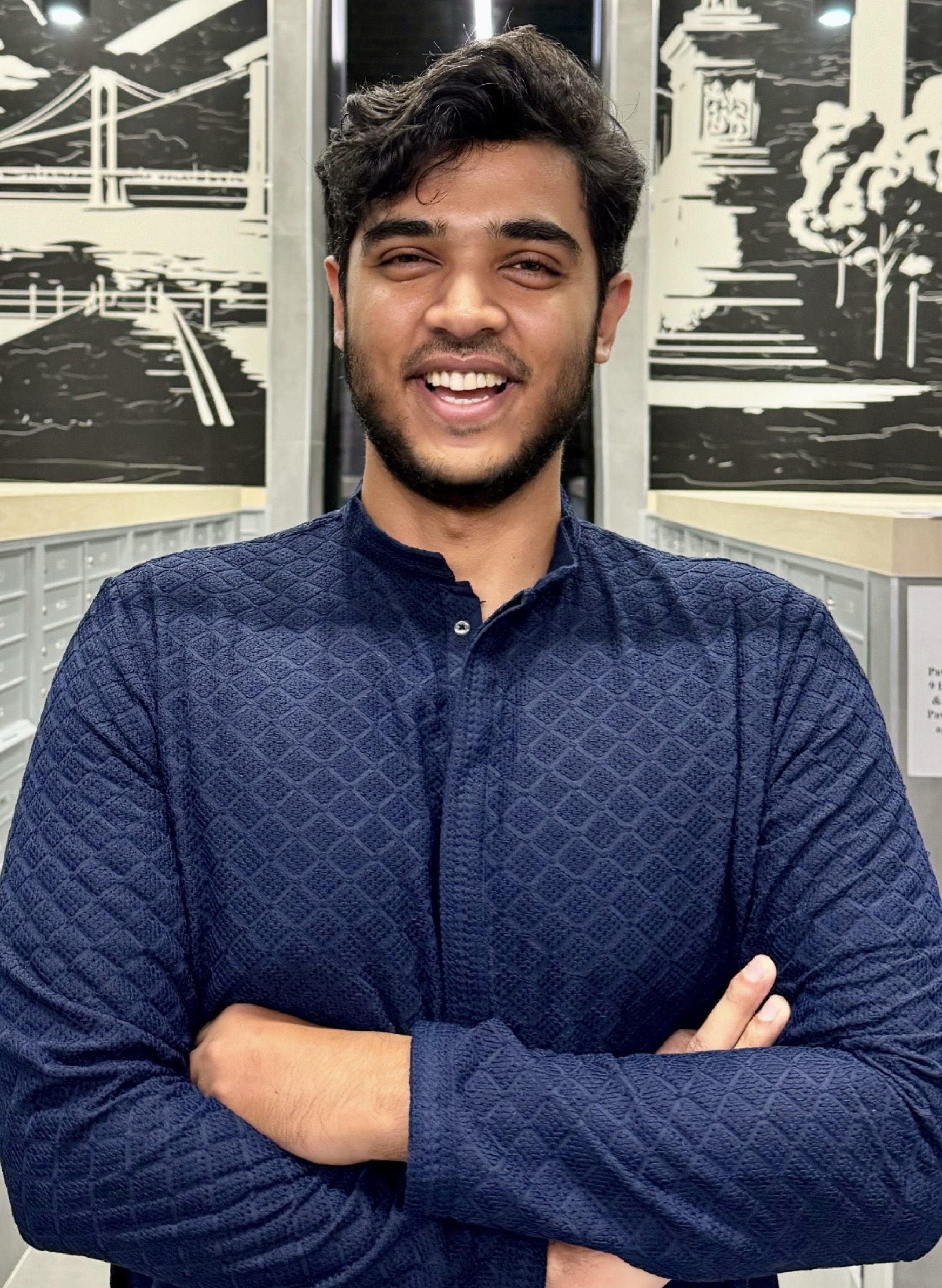}}] {Shubham Vatsal} received the M.S. degree in Computer Science from New York University, New York, USA, in 2023. He is currently a Research Data Scientist in the healthcare domain based in New York, where he applies Machine Learning and Natural Language Processing (NLP) to real-world clinical and operational problems. 

From 2017 to 2021, he was with Samsung Research, focusing on Natural Language Processing (NLP) and Machine Learning. He subsequently worked at IBM Research on Multimodal learning. He has also contributed to a couple of startups, helping build data-driven products using Machine learning and Deep learning. His research interests include Large Language Models (LLMs), Agentic AI, applications of LLMs in healthcare and NLP, Deep Learning and Machine Learning. 

\end{IEEEbiography}


\begin{IEEEbiography}[{\includegraphics[width=1in,height=1.25in,clip,keepaspectratio]{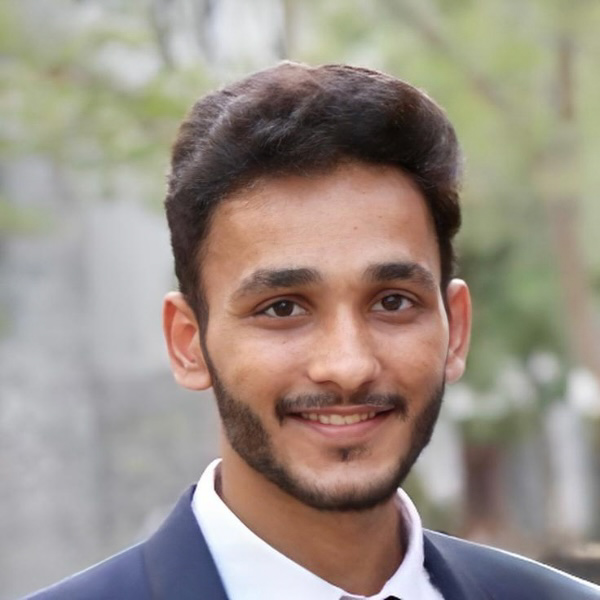}}]{Harsh Dubey} received the M.S. degree in
computer science from New York University,
Courant Institute of Mathematical Sciences, New
York, USA, in 2023. He is currently based in
New York, where he focuses on agentic plat-
forms, multi-agent frameworks, context memory
management, and large-scale workflow execution.
His work centers on AI agents, large language
models (LLMs), retrieval and memory systems,
and the design of scalable, distributed systems
for enterprise applications. From 2018 to 2020, he held engineering positions at Octro and IndusOS, where he worked on backend systems, personalization, and data-driven product development. His professional experience spans machine learning, distributed infrastructure, and applied artificial intelligence. His research interests include agentic AI, large language models, retrieval-augmented generation, computational cognitive modeling, machine learning, and distributed systems
\end{IEEEbiography}

\begin{IEEEbiography}[{\includegraphics[width=1in,height=1.25in,clip,keepaspectratio]{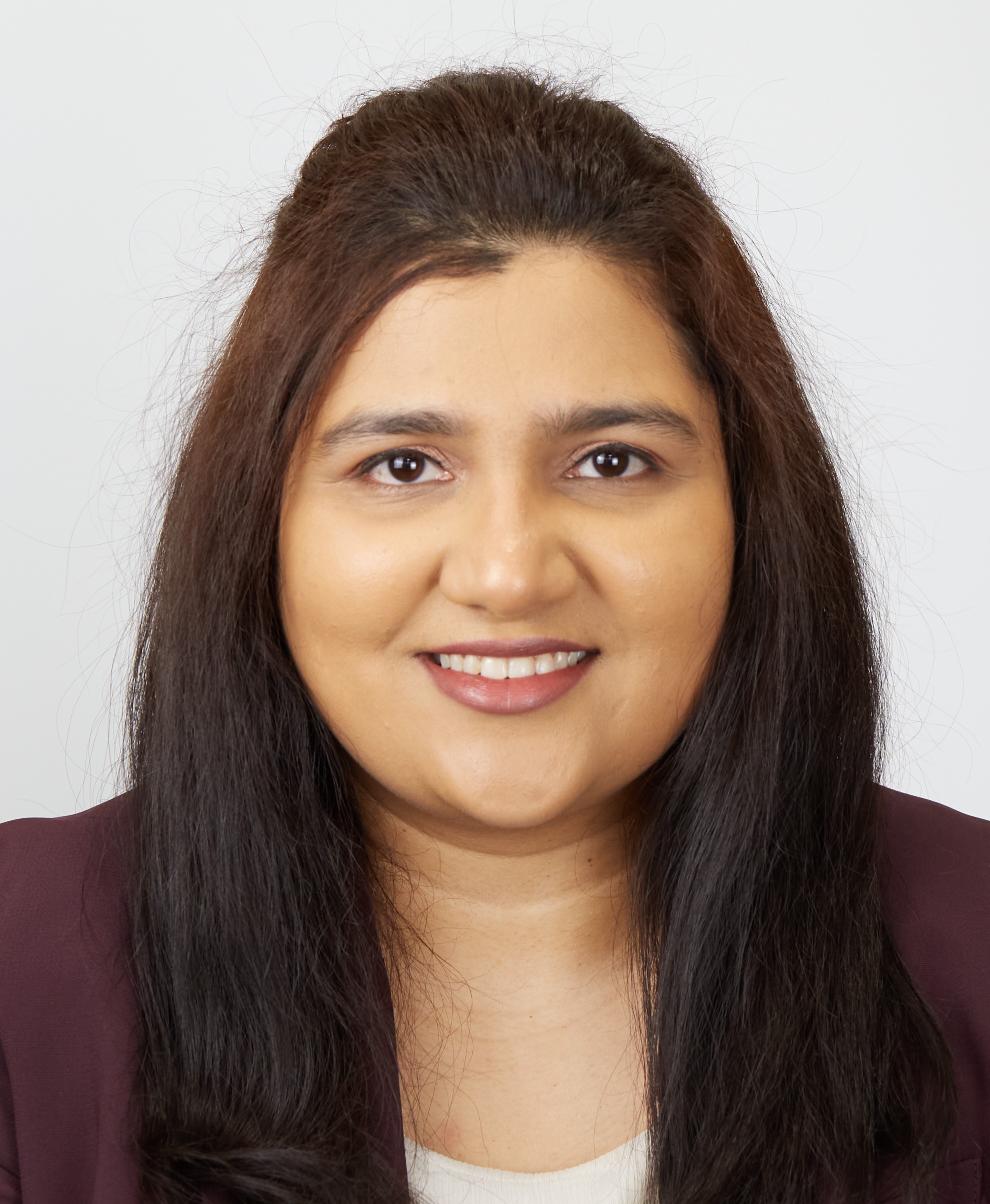}}]{Aditi Singh}
received the M.S. degree in Computer Science from Kent State University, Kent, OH, USA, in 2017, and the Ph.D. degree in Computer Science from Kent State University in 2023. She is an Assistant College Lecturer with the Department of Computer Science, Cleveland State University, Cleveland, OH, USA. Previously, she served as a Graduate Assistant at Kent State University (2017–2023). Her research interests include artificial intelligence and large language models, with emphasis on agentic systems, retrieval and memory for LLMs, text-to-SQL, the Model Context Protocol (MCP), and computer vision for human–robot interaction. She has authored over 30 publications (with several in press) and has received multiple recognitions, including Best Presenter Awards at IEEE CCWC (2023–2025), a Best Paper Award (CCWC 2025), and selection as a Finalist for the Women in AI (Research) Award in 2025. She serves on program committees for venues such as AIET, AIED, EDM, and ICR and reviews for international conferences and journals, including NeurIPS and \textit{ACM Computing Surveys}. She is a Senior Member of the IEEE and chairs the Cleveland ACM-W professional chapter.
\end{IEEEbiography}

\EOD
\end{document}